# A Novel Framework for Significant Wave Height Prediction based on Adaptive Feature Extraction Time-Frequency Network

Jianxin Zhang[1], Lianzi Jiang[1], Xinyu Han[*,2,3], Xiangrong Wang[1]

[1] Shandong University of Science and Technology, College of Mathematics and Systems Science, Qingdao, 266590, China

[2] Shandong University of Science and Technology, College of Civil Engineering and Architecture, Qingdao, 266590, China

[3.] Qingdao Key Laboratory of Marine Civil Engineering Materials and Structures, Qingdao, 266590, China

## Abstract

Precise forecasting of significant wave height ($H_s$) is essential for the development and utilization of wave energy. The challenges in predicting $H_s$ arise from its non-linear and non-stationary characteristics. The combination of decomposition preprocessing and machine learning models have demonstrated significant effectiveness in $H_s$ prediction by extracting data features. However, decomposing the unknown data in the test set can lead to data leakage issues. To simultaneously achieve data feature extraction and prevent data leakage, a novel Adaptive Feature Extraction Time-Frequency Network (AFE-TFNet) is proposed to improve prediction accuracy and stability. It is encoder-decoder rolling framework. The encoder consists of two stages: feature extraction and feature fusion. In the feature extraction stage, global and local frequency domain features are extracted by combining Wavelet Transform (WT) and Fourier Transform (FT), and multi-scale frequency analysis is performed using Inception blocks. In the feature fusion stage, time-domain and frequency-domain features are integrated through dominant harmonic sequence energy weighting (DHSEW). The decoder employed an advanced long short-term memory (LSTM) model. Hourly measured wind speed ($W_s$), dominant wave period ($DPD$), average wave period ($APD$) and $H_s$ from three stations are used as the dataset, and the four metrics are employed to evaluate the forecasting performance. Results show that AFE-TFNet significantly outperforms benchmark methods in terms of prediction accuracy. Feature extraction can significantly improve the prediction accuracy. DHSEW has substantially increased the accuracy of medium-term to long-term forecasting. The prediction accuracy of AFE-TFNet does not demonstrate significant variability with changes of rolling time window size. Overall, AFE-TFNet shows strong potential for handling complex signal forecasting.

## 1. Introduction

### 1.1. Research background

Fossil fuels remain a dominant source of energy worldwide, but they are also one of the largest contributors to carbon emissions, driving global warming and climate change. To achieve the global goals of reducing greenhouse gas emissions and transitioning toward sustainable energy systems, renewable energy sources are critical. Among these, ocean energy has emerged as a particularly promising option due to its zero-carbon emission characteristics. Specifically, wave energy, which harnesses the kinetic energy of ocean waves, holds immense potential as a high-energy renewable resource. Wave energy devices convert the heaving and pitching motion of wave surfaces into electrical power, making use of the ocean's immense, continuously replenished energy reservoir [1].

Significant wave height ($H_s$), defined as the average height of the highest one-third of waves over a given period, serves as a critical parameter for assessing wave energy potential. Specifically, sites with consistently high $H_s$ values offer promising energy yields; however, the variability in wave conditions can compromise investment viability if not accurately predicted. Thus, precise $H_s$ forecasts are vital for optimizing wave energy converter deployment and operation, maximizing energy capture while minimizing downtime during extreme wave events [2]. Beyond wave energy, accurate $H_s$ forecasts also enhance safety and efficiency in diverse marine contexts. Reliable forecasting enables proactive operational planning, mitigating risks associated with adverse sea states and optimizing operational windows, thereby yielding substantial cost savings and enhanced safety protocols [3]. Effective wave forecasting systems contribute critically to maritime safety by providing vital information for ship routing, navigation, and search and rescue operations, thereby safeguarding lives and assets at sea [4].

**1.2. Literature review**

In recent years, many researchers have proposed a variety of $H_s$ forecasting methods, which can be divided into physical numerical models and machine learning (ML) methods. The common wave physical numerical models include the WAM [5, 6], WAVEWATCH III [7, 8], SWAN [9, 10], MIKE21 SW [11, 12], and FVCOM [13, 14], which simulate the temporal and spatial evolution of waves. These models, based on hydrodynamic principles, provide valuable insights into wave propagation and transformation. However, they are computationally intensive, require extensive input data, and their accuracy can be limited by the complexity of wave physics and the parameterization of sub-grid scale processes.

On the contrary, statistical models including auto-regressive (AR) [15], autoregressive integrated moving average (ARIMA) [16], multiple linear regression (MLR), and gray model (GM). While these models are simple in structure and well-defined, making them effective at capturing historical patterns, they exhibit significant limitations in addressing nonlinear and nonstationary data. These characteristics are particularly prevalent in real-world applications, where wave height predictions are often constrained by the limitations of traditional statistical models. Therefore, developing methods capable of effectively handling the nonlinear and nonstationary nature of significant wave height remains a crucial research challenge.

The increasing availability of reliable public databases has driven the development of advanced machine learning models, which are better suited for handling non-linearity and non-stationarity. Compared with numerical models, machine learning models are more computationally efficient, do not require additional input information, and operators do not need to have professional meteorological knowledge to conduct model training and forecasting. Machine learning models, particularly deep learning models, possess powerful learning and adaptive capabilities. Through continuous training and adjustment, these models can maintain high performance when confronted with new data or changing environments and quickly adapt to these changes. Consequently, plenty of ML models have been developed for $H_s$ prediction, such as artificial neural networks (ANN) [17], support vector machines (SVM) [18], decision tree (DT) [19], echo state network (ESN) [20], PatchTST [21], and so on. Despite their successes, existing machine learning models for $H_s$ prediction still face several challenges. Key issues include the generalization capability of models, their sensitivity to overfitting and local optima, and the complexity of hyperparameter tuning. Furthermore, effectively extracting crucial features from complex wave data and constructing machine learning models that fully leverage these features remain critical problems to be addressed.

A critical question emerges: with the support of ML, can data-driven models effectively replace physical numerical models? This needs to be considered from the perspective of computational time and accuracy. Browne et al. [22] gave a comprehensive explanation of using ANNs to estimate waves nearshore and compared results with SWAN model. Results show that ANNs outperformed the physical model in simulations for seventeen nearshore locations around the continent of Australia over a 7-month period. Shamshirband et al. [23] compared SWAN numerical model and machine learning models, the results show that the SWAN model is slightly more precise than machine learning models and the errors of artificial neural networks, extreme learning and support vector machines are comparable. Peach et al. [24] compared three NN models and SWAN models. The results show that the SWAN model and the other three NN models have comparable accuracy. However, the runtime of the SWAN model is about 8000 times that of the NN model. These findings demonstrate that ML methods can achieve precision comparable to numerical models. Optimizing the architecture and parameters of ML models may achieve higher accuracy. Once the ML model is trained, its computational efficiency for making predictions is significantly better than that of numerical models.

Although machine learning models perform better than statistical models in handling nonlinear data, they still face issues such as overfitting, local optima, and sensitivity to hyper-parameters [25]. Moreover, the dramatic fluctuations on $H_s$ remain a challenge for single machine learning models. To address these problems, several hybrid models have been developed to enhance forecasting accuracy. Depending on the type of hybrid architecture, hybrid models can be classified into three categories:

**Single ML model + parameter optimization algorithm**. This type of hybrid model includes a basic machine learning model (such as SVM, DT, and ANN) combined with an optimization algorithm, such as, covariance-weighted least squares (CWLS) algorithm [26], genetic algorithm (GA) [27], Bayesian optimization (BO) [28, 29], Particle swarm optimization (PSO) [30]. For complex datasets and nonlinear problems, optimization algorithms can help models find more refined parameter settings, thereby improving model performance. By optimizing hyperparameters through appropriate search strategies, training time and resource consumption can be reduced, avoiding overfitting or underfitting on new data, and thereby improving the accuracy and generalization ability of the model.

**Time series decomposition preprocessing + single ML model.** In this type of hybrid model, the commonly used neural network model is the long short-term memory (LSTM) recurrent neural network [31]. LSTM possesses a robust ability to learn long-term dependencies, enabling it to utilize historical information for precise predictions. A key innovation of the LSTM architecture lies in the introduction of the "forget gate" a mechanism that optimizes traditional recurrent neural networks. This aims to effectively mitigate the vanishing gradient problem, thereby enhancing the model's performance when processing long time-series data. The purpose of time series decomposition techniques is to separate different model information within a time series, and then to individually forecast each decomposed series. This approach helps reduce the nonlinearity and nonstationary of the time series, thereby decreasing the complexity of the prediction process. wavelet transform (WT) [32], empirical orthogonal function (EOF) [33], seasonal-trend decomposition based on loess (STL) [34], singular spectrum analysis (SSA) [35], variational mode decomposition (VMD) [36], empirical mode decomposition (EMD) [37], and some variants of EMD methods [38], such as, complete ensemble empirical mode decomposition with adaptive noise (CEEMDAN) [39], and improved complete ensemble empirical mode decomposition with adaptive noise (ICEEMDAN) [40] have been employed to predict wave height.

**Ensemble ML system.** As numerous ML models have been proposed, each model has its own limitations. Therefore, it is worth considering whether it is possible to build an ensemble system based on multiple ML models to improve predictive performance and stability. Ahmed et al. [41] developed a hybrid model combining convolutional neural network-long short-term memory-bidirectional gated recurrent unit forecast system (CLSTM-BiGRU), which can simultaneously handle spatiotemporal dependencies, making wave height prediction more accurate. Kumar et al. [42] developed Growing and Pruning Radial Basis Function (GAP-RBF) network, which predictive performance is superior to that of individual models. Patanè et al. [43] proposed the multi-block CNN-LSTM. Its performance in wave height prediction surpasses that of the LSTM model. Hu et al. [19] combined the LSTM and Extreme Gradient Boosting (XGBoost) to predict wave height. The study shows that this integrated hybrid model significantly improves the accuracy of wave height predictions compared to WAVEWATCH III. Wang et al. [44] proposed CNN-BiLSTM-Attention model and trained by WAVEWATCH III reanalysis data. Ikram et al. [27] proposed hybrid neuro-fuzzy model by combining the neuro-fuzzy approach with the different optimization algorithm to predict short term wave height. However, this type of ensemble neural network model has a long training time, as it involves training several networks, making the model overly complex.

Hybrid models, while demonstrating considerable potential in improving $H_s$ forecasting accuracy, present several notable challenges. The incorporation of multiple methodologies inherently increases model complexity and computational overhead. A critical limitation of traditional hybrid approaches, particularly those combining time series decomposition with machine learning techniques, lies in their susceptibility to data leakage. Specifically, when decomposition is applied to the entire dataset (encompassing both training and testing sets), information from the test set may inadvertently influence the training process.

This contamination typically results in overly optimistic performance metrics and compromised generalization capability when the models are deployed in practical applications. Consequently, the development of hybrid models that simultaneously achieve high accuracy, robust performance, and effectively mitigate data leakage remains a crucial research imperative.

Research indicates that models with information leakage exhibit higher accuracy [45]. Some ML models show a high sensitivity to data leakage, and small datasets exacerbate the impact of data leakage [46]. After correcting data leakage issues, the performance of ML models does not show a significant improvement compared to simple regression models [47]. To retain the advantages of data decomposition and address the issue of information leakage, some studies have begun to explore the use of rolling decomposition instead of direct decomposition. Ding et al. [48] proposed a VMD-LSTM-rolling model to predict $H_s$. First, the time series TS1 is generated using the first $k$ data points, followed by VMD and prediction with LSTM. Subsequently, the first point in TS1 is removed, and the data point at time $k+1$ is appended to the end, forming a new time series TS2. TS2 is then subjected to VMD decomposition and LSTM prediction. This rolling update of the time series continues, guiding the prediction across the entire test set. The idea of this rolling model has also been successfully applied to the prediction of other time series, such as wind speed [49], ammonia nitrogen [50], and so on. Nevertheless, existing rolling decomposition models exhibit limitations, particularly in their feature extraction capabilities. A critical research gap exists in developing sophisticated feature extraction methodologies that can be seamlessly integrated into rolling frameworks to capture the intricate time-frequency characteristics of $H_s$ comprehensively. Moreover, a fundamental challenge in this domain lies in establishing a robust predictive framework that simultaneously satisfies multiple critical requirements: maintaining data integrity through prevention of leakage, adaptively capturing the temporal dynamics of significant wave height, and efficiently extracting salient features from complex wave patterns.

To systematically evaluate the existing literature and elucidate these challenges, Table 1 provides a detailed comparison of the reviewed models, highlighting their categories, representative models, main advantages, and disadvantages. The identified limitations underscore the need for an innovative approach that simultaneously addresses multiple critical requirements: effective handling of non-linear and non-stationary wave patterns, prevention of data leakage in rolling forecasts, and robust extraction of informative time-frequency features. To address these fundamental limitations in existing wave height forecasting methodologies, this study proposes a novel Adaptive Feature Extraction Time-Frequency Network (AFE-TFNet) model aimed at maximizing the utilization of observed data and fully leveraging the potential of predictive variables to enhance short- and long-term wave height forecasting performance. The model takes wind speed ($W_s$), dominant wave period ($DPD$), average wave period ($APD$), and $H_s$ as input and uses rolling window segmentation to decompose the entire time series into several subsequences. WT is used to extract local frequency features from the subsequences, while Fast Fourier Transform (FFT) is employed to extract periodic features from the subsequences. The FFT separated sequences are reshaped and convolved with the WT results, integrating multi-scale information to obtain the final feature-extracted sequence. In this model, the feature extraction from different levels are typically integrated with the original input sequence through dominant harmonic sequence energy weighting (DHSEW), ensuring that the model can perform a comprehensive analysis across multiple scales. An LSTM model and the Adaptive moment estimation (ADAM) algorithm [51] are introduced to implement rolling predictions.

### 1.3. Main contributions of the study

The contributions of this work are summarized as follows:

(1) A novel parallel feature extraction stage within a rolling framework has been introduced to overcome the limitations of single transformations in capturing comprehensive frequency-domain information. WT for local frequency feature extraction and FFT for global periodic feature extraction have been innovatively combined from rolling subsequences, strictly within each rolling training window to prevent data leakage.

(2) To comprehensively capture the complex frequency characteristics of wave signals, a novel Frequency Inception

Block (FIB) has been proposed to address multi-scale frequency characteristics. This component employs multiple convolutional kernels of varying sizes to extract features across different frequency ranges from wavelet and reshaped Fourier components. This multi-scale approach effectively captures both high- and low-frequency information, significantly reducing information loss compared to traditional frequency analysis methods.

(3) A novel Dominant Harmonic Sequence Energy Weighting (DHSEW) mechanism has been proposed to address the challenges of time-frequency domain feature integration. This mechanism dynamically adjusts feature weights based on input time series periodicity, prioritizing frequency-domain information for strongly periodic signals while emphasizing time-domain features for non-periodic signals. This adaptive approach enables robust forecasting across diverse wave conditions by effectively integrating complementary time-domain and frequency-domain representations.

(4) An advanced architecture for $H_s$ forecasting has been developed, integrating these components with an LSTM. The proposed AFE-TFNet consistently outperforms existing baseline models in $H_s$ forecasting, as demonstrated across three distinct real-world wave datasets. Extensive experimental evaluations, including comparisons with multiple baseline models and comprehensive ablation studies, rigorously validate the effectiveness of AFE-TFNet and the contribution of each proposed component.

Table 1 Summary of $H_s$ forecasting models.

| Model Category | Representative models | Main advantages | Main disadvantages |
| --- | --- | --- | --- |
| Physical numerical model | WAM [5, 6], WAVEWATCH III [7, 8], SWAN [9, 10], MIKE21 SW [11, 12], FVCOM [13, 14] | The physical mechanisms are well-defined, enabling the simulation of spatiotemporal evolution. | High computational cost. |
| Statistical model | AR [15], ARIMA [16], MLR, GM | Simple structure and efficient forecasting. | Weak nonlinear handling and limited accuracy fluctuations. |
| Machine learning model | ANN [17], SVM [18], DT [19], ESN [20], PatchTST [21] | Stronger nonlinear handling and adaptive learning capability. | Overfitting susceptibility, poor interpretability, and feature extraction relies on expertise. |
| Hybrid model - parameter optimization | CWLS [26], GA [27], BO [28, 29], PSO [30] | Parameter optimization capability, and performance enhancement. | The algorithm's performance heavily relies on its initial value selection. |
| Hybrid model - decomposition preprocessing | WT [32], EOF [33], STL [34], SSA [35], VMD [36], EMD [37], CEEMDAN [39], ICEEMDAN [40] | Reduced data nonlinearity and improved accuracy. | Data leakage risks. |
| Hybrid model - ensemble learning | CLSTM-BiGRU [41], GAP-RBF [42], CNN-LSTM [43], CNN-BiLSTM-Attention [44] | The integration of multiple model strengths leads to enhanced robustness, leveraging their complementary capabilities to improve performance. | Complex model structure, high training cost, and further reduced interpretability. |
| Rolling decomposition model | VMD-LSTM-Rolling [48] | Avoiding data leakage enhances practical applicability in real-world applications. | Feature extraction methods need improvement. Cannot adaptively identify the time-frequency patterns of significant wave height. |

## 1.4. Organization and structure

The remainder of this paper is organized as follows: Section 2 introduces data description. Section 3 provides details about the architecture of the present model. Section 4 introduces four experiments, and the corresponding analyses. Section 5 draws the conclusions of this study.

## 2. Data description

This study utilizes wave observation data provided by the National Data Buoy Center (NDBC) of the United States (https://www.ndbc.noaa.gov/). To validate the model's performance under different geographical and climatic conditions, three buoy stations were selected: No.41010, located east of Cape Canaveral (28°52'41"N, 78°28'0"W). No.46025, located southwest of Santa Monica (33°45'19"N, 119°2'42"W). and No.46029, located west of the Columbia River Mouth (46°9'48"N, 124°29'12"W), as shown in Fig. 1. The datasets from these buoys, denoted as A, B, and C, include observations of wind speed ($W_s$), dominant wave period ($DPD$), average wave period ($APD$), and significant wave height ($H_s$) from January 1, 2000, to December 31, 2012. $H_s$ was chosen as the target variable for prediction, as it is a crucial parameter for assessing ocean wave energy potential and coastal engineering design. $W_s$, $DPD$, and APD as input features were justified by two key factors: their strong correlation with $H_s$ as demonstrated through Pearson correlation coefficient analysis, and the existence of well-defined physical relationships between these parameters and significant wave height [52, 53]. To improve the model's generalization performance and ensure robust evaluation, each dataset was divided into three subsets: 70% for training, 20% for validation, and 10% for testing, maintaining the temporal order of the data to avoid data leakage and ensure realistic forecasting scenarios. Table 2 summarizes the descriptive statistics of each dataset, including the minimum, maximum, average, standard deviation, and median. These descriptive statistics provide a comprehensive overview of the data characteristics.

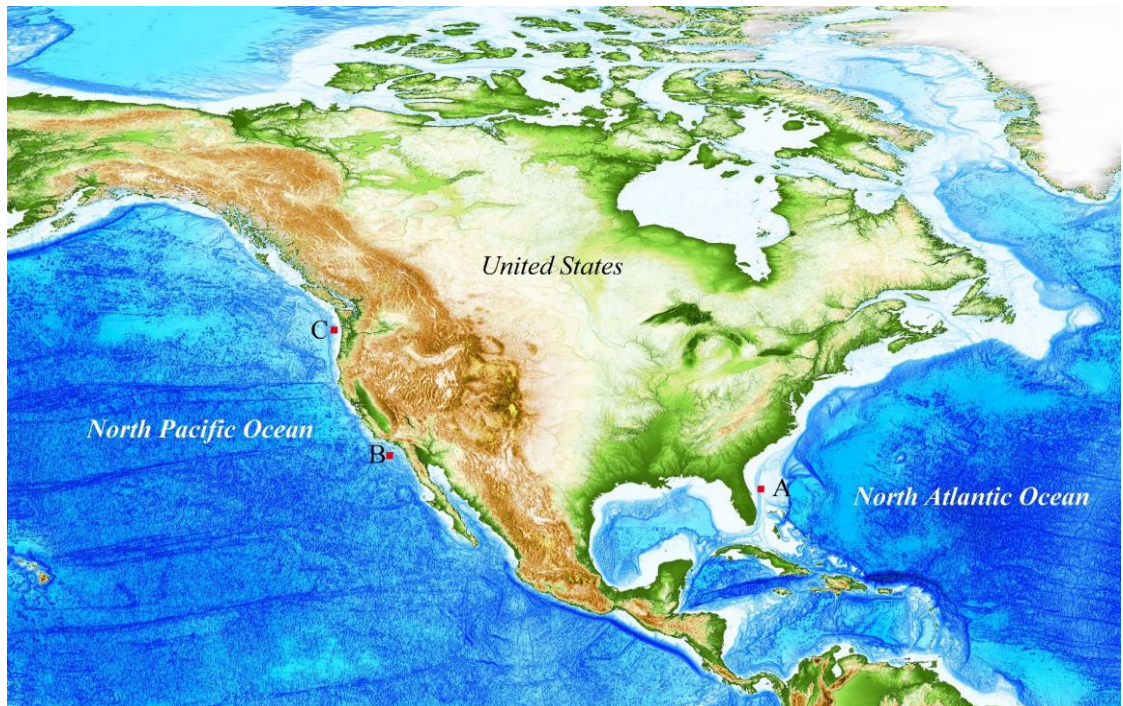

Fig. 1. Buoy locations in this study.

Table 2 Overview of datasets.

| Dataset | Property | All data (100%) | | | | | Dataset | Property | Training (70%) | | | | |
|---|---|---|---|---|---|---|---|---|---|---|---|---|---|
| | | Min | Max | Avg | Std | Median | | | Min | Max | Avg | Std | Median |
| A | $W_s$ (m/s) | 0 | 22.9 | 6.06 | 3.29 | 5.8 | A | $W_s$ (m/s) | 0 | 22.9 | 6.06 | 3.28 | 5.8 |
| | DPD(s) | 2.78 | 25 | 10.73 | 2.94 | 10.81 | | DPD(s) | 2.78 | 25 | 10.72 | 2.94 | 10.81 |
| | APD(s) | 3.23 | 15.56 | 6.98 | 1.46 | 6.83 | | APD(s) | 3.28 | 15.56 | 6.98 | 1.46 | 6.83 |
| | $H_s$ (m) | 0.22 | 13.75 | 2.38 | 1.27 | 2.06 | | $H_s$ (m) | 0.42 | 13.75 | 2.38 | 1.27 | 2.05 |
| B | $W_s$ (m/s) | 0 | 31.4 | 6.14 | 2.97 | 5.9 | B | $W_s$ (m/s) | 0 | 31.4 | 6.13 | 2.96 | 5.9 |
| | DPD(s) | 2.6 | 17.39 | 8.05 | 2.26 | 8.33 | | DPD(s) | 2.6 | 17.39 | 8.05 | 2.27 | 8.33 |
| | APD(s) | 2.86 | 13.06 | 5.43 | 1.01 | 5.29 | | APD(s) | 2.9 | 13.06 | 5.44 | 1.01 | 5.29 |
| | $H_s$ (m) | 0.34 | 10.18 | 1.58 | 0.83 | 1.39 | | $H_s$ (m) | 0.34 | 10.18 | 1.58 | 0.82 | 1.39 |
| C | $W_s$ (m/s) | 0 | 18.4 | 3.58 | 2.39 | 3.1 | C | $W_s$ (m/s) | 0 | 18.4 | 3.57 | 2.39 | 3.1 |
| | DPD(s) | 2.74 | 25 | 12.01 | 3.79 | 12.9 | | DPD(s) | 2.74 | 25 | 12.01 | 3.78 | 12.9 |
| | APD(s) | 3.05 | 14.11 | 6.28 | 1.36 | 6 | | APD(s) | 3.05 | 14.11 | 6.28 | 1.36 | 6 |
| | $H_s$ (m) | 0.34 | 5.39 | 1.14 | 0.45 | 1.04 | | $H_s$ (m) | 0.34 | 5.19 | 1.14 | 0.45 | 1.04 |

| Dataset | Property | Validation (20%) | | | | | Dataset | Property | Testing (10%) | | | | |
|---|---|---|---|---|---|---|---|---|---|---|---|---|---|
| | | Min | Max | Avg | Std | Median | | | Min | Max | Avg | Std | Median |
| A | $W_s$ (m/s) | 0 | 22.7 | 6.06 | 3.32 | 5.8 | A | $W_s$ (m/s) | 0 | 22.6 | 6.09 | 3.33 | 5.8 |
| | DPD(s) | 3.23 | 25 | 10.72 | 2.94 | 10.81 | | DPD(s) | 3.33 | 25 | 10.75 | 2.95 | 10.81 |
| | APD(s) | 3.23 | 14.75 | 6.98 | 1.46 | 6.84 | | APD(s) | 3.57 | 13.4 | 7 | 1.46 | 6.86 |
| | $H_s$ (m) | 0.34 | 9.64 | 2.37 | 1.27 | 2.07 | | $H_s$ (m) | 0.22 | 13.74 | 2.41 | 1.3 | 2.08 |
| B | $W_s$ (m/s) | 0 | 27.2 | 6.15 | 2.99 | 5.9 | B | $W_s$ (m/s) | 0 | 24.4 | 6.11 | 2.97 | 5.9 |
| | DPD(s) | 2.94 | 17.39 | 8.05 | 2.25 | 8.33 | | DPD(s) | 2.6 | 16.67 | 8.03 | 2.24 | 8.33 |
| | APD(s) | 2.91 | 12.55 | 5.43 | 1.02 | 5.3 | | APD(s) | 2.86 | 12.77 | 5.43 | 1 | 5.29 |
| | $H_s$ (m) | 0.35 | 9.63 | 1.58 | 0.84 | 1.39 | | $H_s$ (m) | 0.36 | 8.88 | 1.57 | 0.83 | 1.38 |
| C | $W_s$ (m/s) | 0 | 17.6 | 3.59 | 2.39 | 3.1 | C | $W_s$ (m/s) | 0 | 16.8 | 3.6 | 2.4 | 3.1 |
| | DPD(s) | 3.13 | 25 | 12.01 | 3.81 | 12.9 | | DPD(s) | 2.94 | 25 | 12.05 | 3.78 | 12.9 |
| | APD(s) | 3.28 | 13.99 | 6.27 | 1.37 | 6.01 | | APD(s) | 3.25 | 13.36 | 6.31 | 1.37 | 6 |
| | $H_s$ (m) | 0.34 | 4.58 | 1.14 | 0.45 | 1.04 | | $H_s$ (m) | 0.37 | 4.59 | 1.15 | 0.45 | 1.04 |

## 3. Proposed Method

This section presents a detailed description of the AFE-TFNet framework workflow. Fig. 2 illustrates the proposed AFE-TFNet framework for significant wave height prediction, which adopts an encoder-decoder structure. As depicted in the "Rolling process" panel of Fig. 2, AFE-TFNet employs a rolling window approach to process the time series data and generate forecasts iteratively. The original time series is transformed into multiple overlapping subsequences using a sliding window of a fixed size. The window size, denoted as $T$, defines the length of each subsequence, and the window advances forward one step at a time. For the original time series $X = \{x_1, x_2, ..., x_m\}$, the rolling process generates a series of subsequences:

$$\begin{aligned}
&\textit{Time series 1:} \{x_1, x_2, ..., x_T\} \\
&\textit{Time series 2:} \{x_2, x_3, ..., x_{T+1}\} \\
&\textit{Time series 3:} \{x_3, x_4, ..., x_{T+2}\} \\
&... \\
&\textit{Time series n:} \{x_{m-T+1}, x_{m-T+2}, ..., x_m\}
\end{aligned} \quad (1)$$

The first time series subsequence, denoted as Time Series *1*, is constructed using the initial $T$ data points. Subsequent subsequences, such as Time Series *i*, are generated by shifting the window forward one time step. For each resulting subsequence, the AFE-TFNet encoder processes the data to extract relevant features, which the decoder then uses to produce a forecast for the next time step.

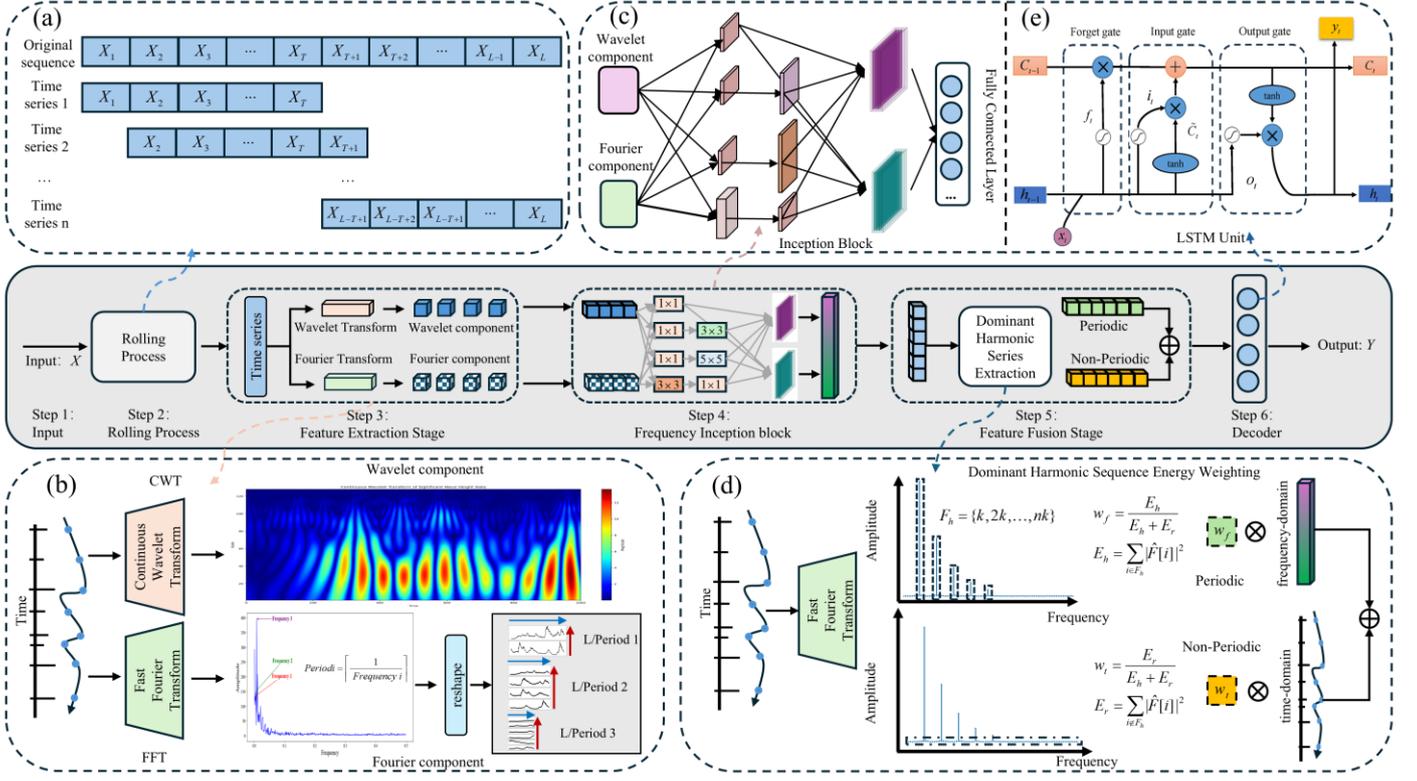

Fig. 2. Overall architecture of AFE-TFNet.

This rolling approach ensures that the model relies solely on past and present information for forecasting, preventing future information leakage during feature extraction and training while preserving the causality of time series prediction. Importantly, the rolling window segmentation is applied independently to the training, validation, and testing datasets following the initial dataset split. This ensures that the rolling window process is confined to each dataset split, with no data mixing or information exchange across datasets. Such strict separation further guarantees the absence of data leakage throughout the model development and evaluation process.

The encoder has two stages: feature extraction and feature fusion, as shown in Algorithm 1. In the feature extraction stage, WT and FT are combined to extract local and global frequency-domain features from the input signals. Additionally, the Frequency Inception block is applied to perform multi-scale analysis on the extracted frequency-domain features. After feature extraction, the feature fusion stage utilizes a DHSEW mechanism to integrate time- and frequency-domain features, thereby adapting to different signal variation patterns. The decoder employs an LSTM network to process and analyze the fused multi-dimensional features. The LSTM network captures the temporal dependencies between features, producing accurate wave height predictions.

```
Algorithm 1 The Process Flow of AFE-TFNet
Require: Wave observation series X = {x_1, x_2, ..., x_t}
Ensure: Predict ŷ_{t+l}
// Encoder
// 1. Feature Extraction Stage
 1: X_wavelet = Wavelet Transform(X)
 2: X_fourier = Reshape(Fourier Transform(X))
 3: X_inception = Inception Block(X_wavelet, X_fourier)
 4: X_fre = Fully Connected Layer(X_inception)
// 2. Feature Fusion Stage
 5: E_w, E_f = Dominant Harmonic Series Extraction(X)
 6: w_f = E_f / (E_w + E_f),  w_t = E_w / (E_w + E_f)
 7: X_fused = w_f × X_fre + w_t × X
// Decoder
 8: h_t = LSTM(X_fused)
 9: ŷ_{t+l} = Output Gate(h_t)
```

Algorithm. 1. The process flow of AFE-TFNet.

## 3.1. Feature extraction stage

Accurate extraction of frequency-domain features is crucial for the performance of significant wave height prediction models. However, due to the complex frequency characteristics and multi-scale information in wave observation signals, a single transformation (such as WT or FT) fails to effectively capture both global and local frequency-domain information [54]. To address this, this study proposes a parallel feature extraction method that combines WT and FT to more comprehensively capture frequency-domain features, as illustrated in the "Feature Extraction Stage" panel of Fig. 2. This hybrid approach enables the model to overcome the limitations of single transformations and provides a more robust representation of the signal's frequency content. Additionally, this study introduces an Inception block that performs multi-scale convolution operations to further extract both high- and low-frequency information at different scales, effectively reducing information loss and enhancing model performance.

### 3.1.1. Wavelet Transform

Due to the non-stationary nature of wave data, the frequency components vary over time, making it crucial to effectively extract local frequency-domain features for improving prediction accuracy. Wavelet Transform (WT), especially Continuous Wavelet Transform (CWT), can locally decompose the signal in both the time and frequency domains, capturing instantaneous features and frequency variations [55].

Consider a time-series input signal $X = [x_0,...,x_{T-1}] \in \mathbb{R}^{T \times d}$, where $T$ is the length of the time series and $d$ denotes the dimensionality of each time step. The CWT is defined as follows:

$$CWT(a,b) = \int x_t \cdot \psi^* \left( \frac{t-b}{a} \right) dt \tag{2}$$

where $\psi^*$ is the complex conjugate of the mother wavelet function, $a$ is the scale factor, and $b$ is the translation factor.

To advance the accurate prediction of rapidly changing signals and improve model performance, this study selected the Morlet wavelet, recognized for its superior time-frequency localization. In contrast to other basis functions, the Morlet wavelet is particularly effective at capturing local signal features, significantly reducing information loss [56]. In terms of scale selection, this study employed logarithmically spaced scale values within the range $[2^{-1}, 2^5]$ to ensure the effective extraction of local features at multiple resolutions.

### 3.1.2. Fourier Transform

Focusing exclusively on local features may result in overlooking critical global trends, leading to less accurate predictions. To address this, this study applies the Fourier Transform (FT), a signal processing technique that extracts global frequency-domain features, decomposing signals into sine wave components and providing essential global spectral information. By employing the Fast Fourier Transform (FFT), computational complexity is significantly reduced. The discrete form of the FFT is given by:

$$F(k) = \sum_{t=0}^{T-1} x(t) \cdot e^{-i2\pi kt/T} \tag{3}$$

where $F(k)$ denotes the Fourier coefficient at the discrete frequency $k$, and $T$ represents the signal length.

We propose a novel frequency reshape operation to capture periodic patterns at different scales, as illustrated in Fig. 3. While Fig. 3 demonstrates this process using $H_s$ as an example, the frequency reshape operation is implemented across all input variables, including $W_s$, *DPD*, and *APD*. This comprehensive transformation enables the extraction of periodic features from each variable, thereby providing robust frequency domain information essential for subsequent feature fusion and model training. The process begins by applying FFT to the input time series, generating complex-valued Fourier coefficients *F(i)*, which represent the signal's frequency components. The amplitude spectrum $A = [a_1, \ldots, a_T]$ is then derived by calculating the magnitude of each Fourier coefficient $a_i = |F(i)|$. The top $k$ dominant frequencies $\{f_1,...,f_k\}$ are identified by selecting those with the highest amplitudes in the spectrum. For each selected frequency $f_i$, its corresponding period $P_i = \lceil 1/f_i \rceil$ is calculated, with the floor function applied to ensure integer values. The original signal is subsequently segmented based on these periods, with zero-padding applied as needed to ensure uniform segment lengths. Finally, these segments are reshaped into two-dimensional signal maps $\mathbf{X}^i_{fourier}$ of size $P_i \times T/P_i$ as follows:

$$\mathbf{X}^i_{fourier} = \text{Reshape}_{P_i}\left(\text{Padding}(X)\right), i \in \{1,\cdots,k\} \tag{4}$$

where Padding(·) denotes the zero-padding applied to the time series, ensuring the padded data is divisible by the period $P_i$. Subsequently, $\text{Reshape}_{P_i}$ converts the zero-padded signal into a two-dimensional matrix of size $P_i \times T/P_i$. A set of two-dimensional matrices $\{X^1_{fourier},\ldots,X^k_{fourier}\}$ is then constructed based on the selected frequencies and their corresponding periods. This reorganization provides an effective representation, facilitating the analysis of frequency variations in a two-dimensional space.

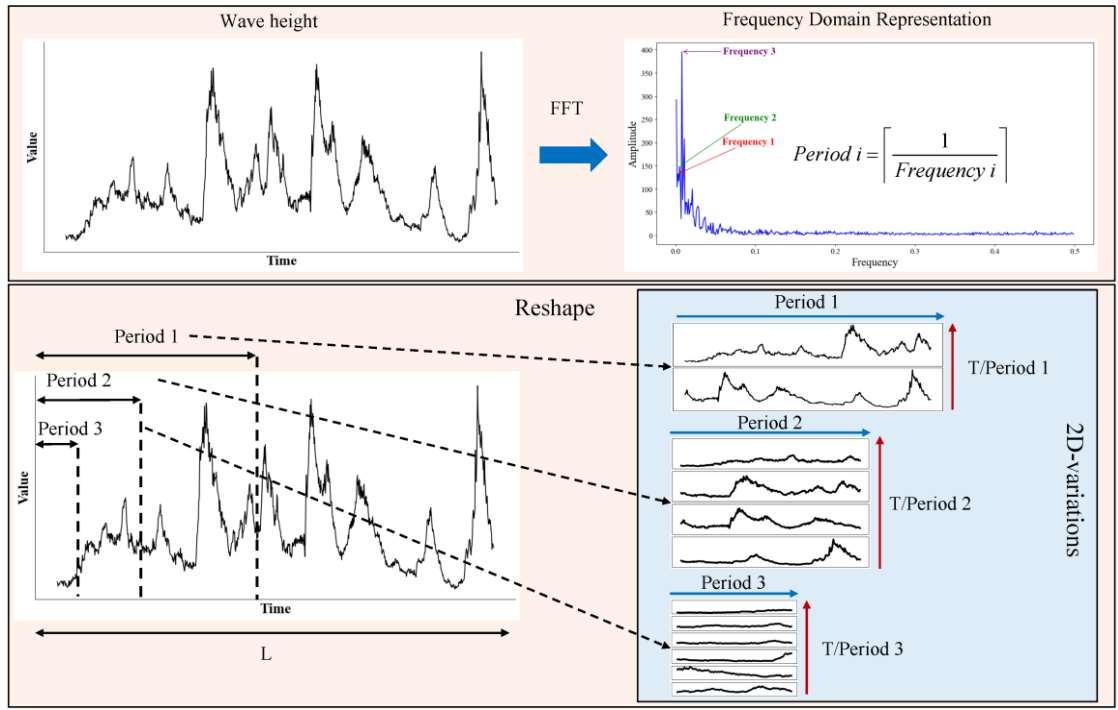

Fig. 3. An example illustrating the frequency reshape process.

**3.1.3. Frequency Inception block**

Accurately capturing frequency characteristics in complex signals is essential for wave height prediction. However, conventional frequency-domain transformations often overlook critical details, focusing solely on either local or global features. To overcome this limitation, this study proposes the Frequency Inception Block (FIB), inspired by the work of Szegedy et al. [57], which integrates frequency domain analysis with multi-scale convolution operations.

Table 3 Detailed architecture of frequency inception block.

| Layer type | Kernel size | Stride | Padding | Filters | Activation |
|---|---|---|---|---|---|
| 1×1 Convolution | 1×1 | 1×1 | Valid | 64 | ReLU |
| 3×3 Convolution | 3×3 | 1×1 | Same | 128 | ReLU |
| 5×5 Convolution | 5×5 | 1×1 | Same | 128 | ReLU |
| 3×3 Max Pooling | 3×3 | 1×1 | Same | - | ReLU |

The Frequency Inception block (FIB), illustrated in the "Inception block" panel of Fig. 2, employs multi-scale convolutional filters to capture WT and reshaped FFT features across different frequency bands. The detailed architecture of the FIB is presented in Table 3. The block begins with three parallel 1×1 convolutional layers, each configured with 64 filters,

1×1 kernel size, 1×1 stride, and ReLU activation function. These 1×1 convolutions serve dual purposes: dimensionality reduction to decrease computational complexity, and preliminary feature extraction to capture initial compact representations from the concatenated wavelet and Fourier features. The architecture then branches into two parallel paths utilizing 3×3 and 5×5 convolution kernels to capture high and low-frequency features, respectively. The 3×3 convolutional layer, configured with 128 filters, 1×1 stride, and "same" padding, effectively captures finer-grained features and higher frequency components. In parallel, the 5×5 convolutional layer, also equipped with 128 filters, 1×1 stride, and "same" padding, focuses on broader patterns and lower frequency components due to its larger receptive field. Both layers utilize ReLU activation functions. A 3×3 max pooling operation with 1×1 stride is implemented to extract prominent local information. This operation not only reduces the spatial dimensions of feature maps but also enhances the model's translation invariance by selecting maximum values within each 3×3 region. The final stage concatenates features from different scales and processes them through a fully connected layer, achieving comprehensive frequency information capture.

Compared to traditional methods, the FIB significantly enhances the model's ability to capture both intra-cycle and inter-cycle variation features. Intra-cycle features, which represent local period $P_i$ details, are derived from the reshaped FFT. Inter-cycle variations, reflecting changes in frequency components over time, are extracted from 2D time-frequency maps generated by WT. By processing features at multiple scales and periods in parallel, the FIB overcomes the limitations of single analysis. This ensures that the model can fully capture and analyze the multi-frequency and multi-cycle variations in complex wave signals.

The outputs from both the WT and the reshaped FFT features, after processing through the FIB, are fused via a fully connected layer as follows:

$$X_{fre} = W(X_{fft}, X_{wt}) + b \tag{5}$$

where $W$ represents the weight.

### 3.2. Feature fusion stage

In predicting significant wave height, time-domain features capture dynamic temporal patterns, while frequency-domain features reveal periodic frequency components. However, existing models often struggle to capture both dynamic temporal patterns and periodic frequency components of wave data, resulting in less reliable predictions. Inspired by Ye et al. [58], this study proposes a mechanism called the Dominant Harmonic Sequence Energy Weighting (DHSEW), which dynamically adjusts fusion weights by analyzing signal periodicity, as illustrated in the "Feature Fusion Stage" panel of Fig. 2. For signals with strong periodicity, the weight of frequency-domain features is higher, whereas for non-periodic signals, time-domain features are given more weight. As shown in Fig. 4, strongly periodic signals exhibit dominant harmonic components, characterized by prominent peaks at the fundamental frequency and its harmonics in the amplitude-frequency spectrum. The amplitude of these peaks reflects the signal's strength at each frequency, with the energy proportional to the square of the amplitude. In contrast, non-periodic signals do not exhibit significant peaks. Based on this observation, we quantify signal periodicity by calculating the energy ratio of the dominant harmonic components relative to the total signal energy.

The key to identifying these dominant harmonic components is determining the fundamental frequency, which is typically the frequency component with the highest amplitude. Specifically, for the input signal $X = [x_0, ..., x_{T-1}] \in \mathbb{R}^{T \times 1}$, the FFT is applied to obtain its Fourier coefficient, and the component with the highest amplitude is selected as the fundamental frequency, denoted as $F(k)$. Next, the integer multiples of the fundamental frequency, i.e., the harmonic components $\{F(2k), F(3k), ..., F(nk)\}$, are identified, where $n$ denotes the number of harmonics. In this study, five harmonic components are chosen, as previous research in ocean engineering and marine meteorology has shown that these components capture most of the wave energy [59].

The total energy of the fundamental frequency and its harmonics $E_h$ is calculated:

$$E_h = \sum_{i=1}^{n} |F(ik)|^2, \quad (6)$$

where $F(ik)$ denotes the Fourier coefficient of the $i^{th}$ harmonic in the spectrum, $|F(ik)|$ is the magnitude, $k$ is the index of the fundamental frequency, and $n$ is the number of harmonics.

The total energy of the spectrum $E_f$ is calculated as:

$$E_f = \sum_{\Omega=0}^{T-1} |F(\Omega)|^2, \quad (7)$$

where $F(\Omega)$ represents the Fourier coefficient of any frequency component in the spectrum.

The energy ratio of the dominant harmonic sequences is calculated as $w_f = \dfrac{E_h}{E_f}$, where $w_f$ represents the frequency domain weight, and $w_t = 1 - w_f$ represents the time domain weight. These weights are dynamically adjusted based on the signal periodicity.

Finally, the time- and frequency-domain features are dynamically weighted using $w_t$ and $w_f$, respectively, as shown in the following formulas:

$$X_{fused} = w_f X_{fre} + w_t X, \quad (8)$$

where $X$ represents the wave observation data, $X_{fre}$ denotes the frequency-domain features, and $X_{fused}$ denotes the final integrated feature vector.

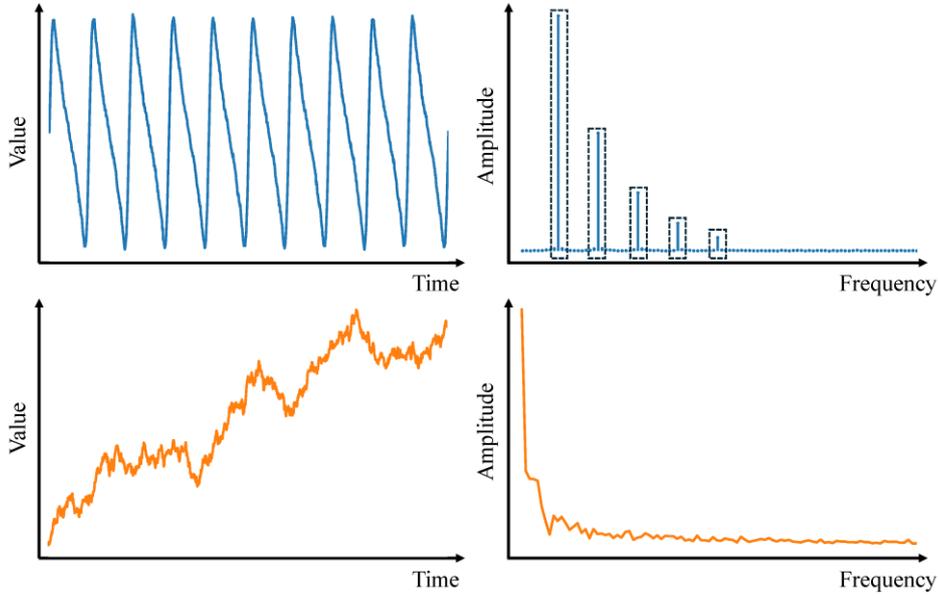

Fig. 4. Harmonic sequence and energy distribution in periodic and non-periodic signals.

### 3.3. Decoder

The LSTM network is used as the decoder to capture long-term dependencies in the wave data. A common LSTM structure is illustrated in Fig. 5. Specifically, at time t, the core of the LSTM consists of the cell state $C_t$, and three gate structures: the input gate $i_t$, the forget gate $f_t$, and the output gate $o_t$. The cell state is updated as follows:

**Step 1:** The forget gate decides which information to ignore from the previous cell state $C_{t-1}$. This process is regulated by the sigmoid activation function, expressed as:

$$f_t = sigmoid\left(W_{fh} h_{t-1} + W_{fx} X_{fused} + b_f\right) \quad (9)$$

where $W_{fh}$ and $W_{fx}$ are weight matrices, and $b_f$ is the bias term. Note that $X_{fused}$ is the weighted combination of time- and frequency-domain representations.

**Step 2:** The input gate controls which information to update the cell state. In this step, the input vector $i_t$ is obtained using the sigmoid function, while the candidate vector $\tilde{C}_t$ is generated through the tanh function:

$$i_t = \text{sigmoid}\left(W_{ih}h_{t-1} + W_{ix}X_{fused} + b_i\right) \quad (10)$$

$$\tilde{C}_t = \tanh\left(W_{ch}h_{t-1} + W_{cx}X_{fused} + b_c\right) \quad (11)$$

where $W_{ih}$, $W_{ix}$, $W_{ch}$ and $W_{cx}$ are weight matrices, and $b_i$, $b_c$ are bias term.

**Step 3:** The cell state $C_t$ is updated based on the forget and input gates, following the equation:

$$C_t = f_t C_{t-1} + i_t \tilde{C}_t \quad (12)$$

**Step 4:** The output gate controls which information is passed to the next step. The hidden state $h_t$ is computed as follows:

$$o_t = \text{sigmoid}\left(W_{oh}h_{t-1} + W_{ox}X_{fused} + b_o\right) \quad (13)$$

$$h_t = o_t \tanh(C_t) \quad (14)$$

where $o_t$ is an activation vector, $W_{oh}, W_{ox}$ are weight matrices, and $b_o$ is the bias term. The hidden state $h_t$ is the final output of this process, which is passed to the next time step.

LSTM networks efficiently capture long-term dependencies in time series data by processing and passing state information over multiple time steps.

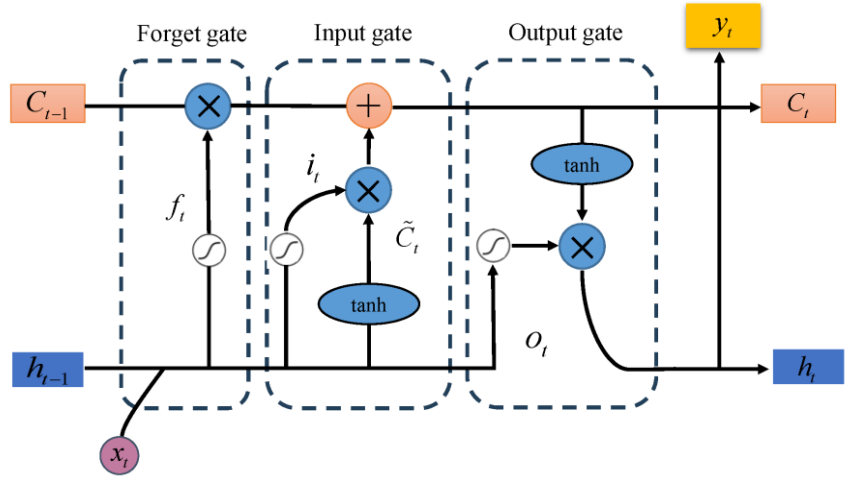

Fig. 5. Diagram showing the structure of an LSTM cell.

### 3.4. Evaluation metrics

In this paper, four common metrics are used to evaluate prediction results: root mean square error (*RMSE*), mean absolute error (*MAE*), mean absolute percentage error (*MAPE*), and correlation coefficient (*R*), each indicating a different aspect of prediction accuracy. The above metrics are defined as:

$$RMSE = \sqrt{\frac{1}{n}\sum_{i=1}^{n}(\hat{y}_i - y_i)^2} \quad (15)$$

$$MAE = \frac{1}{n}\sum_{i=1}^{n}|\hat{y}_i - y_i| \quad (16)$$

$$MAPE = \frac{1}{n}\sum_{i=1}^{n}\left|\frac{\hat{y}_i - y_i}{y_i}\right| \times 100\% \quad (17)$$

$$R = \frac{\sum_{i=1}^{n}(y_i - \bar{y})(\hat{y}_i - \bar{\hat{y}})}{\sqrt{\sum_{i=1}^{n}(y_i - \bar{y})^2 \sum_{i=1}^{n}(\hat{y}_i - \bar{\hat{y}})^2}} \qquad (18)$$

where $\hat{y}_i$ is the $i^{th}$ predicted value, $y_i$ is the $i^{th}$ observed value, $\bar{y}$ and $\bar{\hat{y}}$ represent the mean values of the observed and prediction value, respectively. The total number of data points in the test set is denoted by $n$.

## 4. Experimental results

### 4.1. Baseline models

This section introduces the baseline models used for comparison with our proposed approach. The models chosen for this comparison are NaiveDrift, XGBoost, CatBoost, LightGBM, LSTM, Temporal Convolutional Network (TCN), and Multi-Scale Wavelet Network (MWNet). A brief description follows:

1. Naive Drift: This is a simple yet effective baseline model for time series forecasting, particularly for series with trend. Naive Drift model extrapolates the forecast based on the drift observed in the historical data. Specifically, it calculates the average change between consecutive data points in the training data and projects this drift forward from the last observed value to generate future predictions.

2. XGBoost (Chen and Guestrin [60]): XGBoost is a highly popular and efficient gradient boosting algorithm. It is an ensemble learning method that sequentially builds decision trees, with each tree attempting to correct the errors of its predecessors. XGBoost incorporates regularization techniques to prevent overfitting and is known for its high accuracy and scalability in various machine learning tasks, including regression and time series forecasting.

3. CatBoost (Prokhorenkova et al. [61]): CatBoost is another gradient boosting algorithm that offers several advantages, particularly in handling regression features and reducing overfitting. CatBoost employs ordered boosting and symmetric decision trees, which help to improve accuracy and robustness.

4. LightGBM (Ke et al. [62]): LightGBM is a gradient boosting framework developed by Microsoft, known for its speed and efficiency, especially with large datasets. This is achieved through techniques such as gradient-based one-side sampling and exclusive feature bundling.

5. LSTM (Graves and Alex [63]): LSTM is developed to address the vanishing gradient problem commonly seen in traditional RNNs. By utilizing memory cells and gates, LSTM effectively controls the flow of information, which allows it to capture long-term dependencies in sequential data, making it suitable for time series forecasting.

6. TCN (Bai et al. [64]): TCN is a CNN-based model designed for sequence modeling. It uses dilated convolutions, which allow the network to achieve a large receptive field, effectively modeling long-term dependencies in time series data.

7. MWNet (Wang et al. [65]): MWNet is a neural network that decomposes time series data into various frequency components, facilitating the extraction of multi-scale features for modeling both high- and low-frequency patterns.

### 4.2. Experimental settings

The deep learning models were implemented using the PyTorch framework and trained using ADAM optimization. The hyperparameters were configured as follows: initial learning rate of $1 \times 10^{-3}$, mini-batch size of 32, and maximum training duration of 100 epochs. To mitigate overfitting, we employed two regularization strategies: dropout with a rate of 0.1 in the output layers and early stopping with a patience of 10 epochs on validation loss. All experiments were conducted using a single NVIDIA GeForce RTX 3090 GPU.

### 4.3. Overall performance comparison: Main results

To assess the statistical significance of differences in prediction error distributions between AFE-TFNet and baseline models, we employed the Wilcoxon signed-rank test on their respective prediction error sequences. Error_AFE_TFNet is defined as the *MAE* between the predicted and observed values of AFE-TFNet. Similarly, the other variables are defined as the *MAE* between predicted and observed values of each model. As shown in Table 4, all comparisons yielded negative Z-values, indicating that the comparative models consistently exhibited higher prediction errors than AFE-TFNet. The statistical significance of these differences was confirmed by p-values below the conventional threshold of 0.05, with numerous comparisons achieving highly significant levels ($p < 0.001$). These findings provide robust statistical evidence to reject the null hypothesis of error equivalence in all cases, confirming that the observed performance differences cannot be attributed to random variation. The results of this non-parametric analysis demonstrate conclusively that AFE-TFNet achieves statistically superior prediction performance compared to all benchmark models.

Table 4 Nonparametric tests.

| Nonparametric tests | Z-value | p-value (two-tailed) |
| --- | --- | --- |
| Error_AFE_TFNet - Error_NaiveDrift | -2.103 | .035 |
| Error_AFE_TFNet - Error_XGBoost | -2.605 | .009 |
| Error_AFE_TFNet - Error_CatBoost | -3.193 | .001 |
| Error_AFE_TFNet - Error_LightGBM | -2.589 | .010 |
| Error_AFE_TFNet - Error_LSTM | -49.353 | .000 |
| Error_AFE_TFNet - Error_TCN | -9.816 | .001 |
| Error_AFE_TFNet - Error_MWNet | -42.143 | .000 |

To evaluate the prediction performance of the models in both the short and long term, the experiment considered 1 h, 3 h, 6 h, and 12 h predictions. Fig. 6 shows four statistical metrics for all models across different datasets. As the prediction horizon increases, the performance of all models declines. However, AFE-TFNet consistently shows a significant advantage, with the performance gap relative to the baseline models gradually increasing. Specifically, at the 1 h prediction, the performance of all models is similar, with AFE-TFNet showing a marginal improvement. At the 3 h prediction, while the performance of all models remains comparable, AFE-TFNet maintains a small advantage. As the prediction horizon extends to 6 h and beyond, AFE-TFNet gradually establishes a significant performance gap. Particularly, at the 12 h prediction, the performance disparity between AFE-TFNet and the baseline models becomes the most pronounced. These results suggest that AFE-TFNet consistently outperforms the baselines models in long-term significant wave height predictions. The specific metrics are shown in Tables. 5-7, which confirm the result described above. Specifically, AFE-TFNet achieves an average reduction of 20.23% in *RMSE*, 21.16% in *MAE*, 28.7% in *MAPE*, and a 5.86% improvement in *R* compared to the second-best model.

To visually assess the performance of AFE-TFNet, Figs. 7-10 present the observed and predicted values of all models on three datasets, from May 2010 to December 2012, for the 1 h, 3 h, 6 h, and 12 h predictions. The last month is magnified to better highlight the prediction details. As shown in Fig. 7, all models closely match the observed curves in the 1 h prediction. This indicates strong performance across models for short-term forecasts. In the 3 h prediction, the prediction curves continue to follow the observed values, although slight deviations appear, as shown in Fig. 8. However, as the prediction horizon extends to 6 h prediction, a notable decline in performance is observed across all models, as shown in Fig. 9. The prediction curves show increased fluctuation, reflecting the inherent challenges of mid-term predictions in highly dynamic environments, such as those with rapid and unpredictable changes. Among the models, MWNet shows larger errors in predicting volatility. In contrast, AFE-TFNet demonstrates better performance, maintaining a smaller overall error. In the 12 h long-term prediction, the divergence in model performance becomes more pronounced, particularly in capturing highly fluctuating regions. For instance, in the magnified section of dataset A in Fig. 10, AFE-TFNet demonstrates significantly higher predictive accuracy than the other models in capturing several highly fluctuating patterns. This improvement can be attributed to the AFE-TFNet's feature extraction method, which leverages WT and FFT decomposition to capture critical information from various frequency components, thereby enhancing overall prediction accuracy, particularly in highly fluctuating conditions.

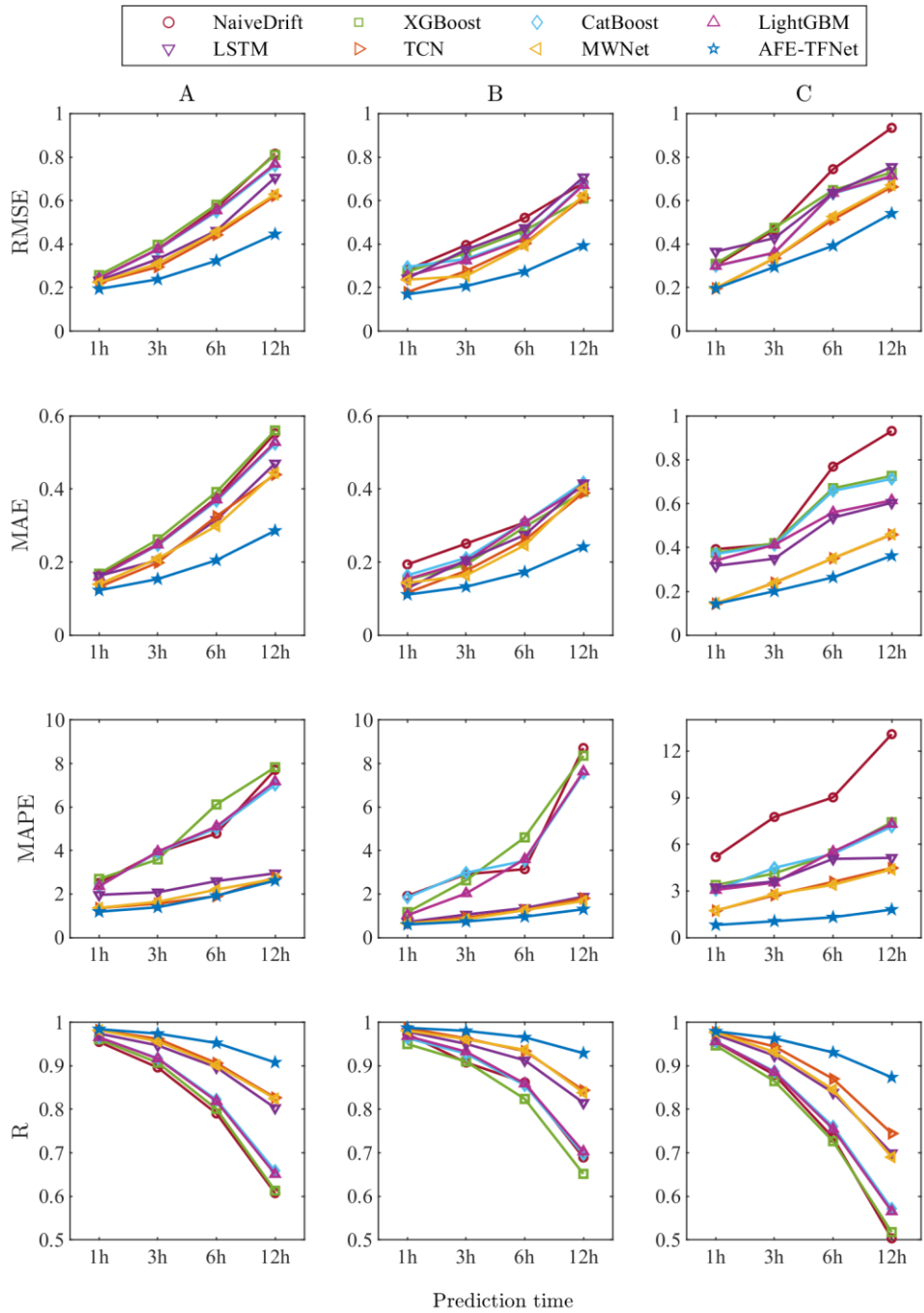

Fig. 6. Performance comparison between AFE-TFNet and baseline models based on *RMSE*, *MAE*, *MAPE*, and *R*.

Table 5 Performance metrics of all models across dataset A.

| Station | Model | Metrics | Prediction time | | | |
|---|---|---|---|---|---|---|
| | | | 1h | 3h | 6h | 12h |
| A | NaiveDrift | RMSE | 0.2437 | 0.3770 | 0.5667 | 0.8156 |
| | | MAE | 0.1642 | 0.2494 | 0.3755 | 0.5521 |
| | | MAPE | 2.5123 | 3.9068 | 4.7859 | 7.7083 |
| | | R | 0.9549 | 0.8961 | 0.7905 | 0.6073 |
| | XGBoost | RMSE | 0.2575 | 0.3967 | 0.5808 | 0.8097 |
| | | MAE | 0.1686 | 0.2621 | 0.3917 | 0.5600 |
| | | MAPE | 2.6977 | 3.5935 | 6.1177 | 7.8290 |
| | | R | 0.9609 | 0.9071 | 0.8009 | 0.6130 |
| | CatBoost | RMSE | 0.2426 | 0.3741 | 0.5487 | 0.7604 |

|  |  | MAE | 0.1596 | 0.2459 | 0.3670 | 0.5233 |
|  |  | MAPE | 2.3688 | 3.9176 | 5.0287 | 7.0193 |
|  |  | R | 0.9653 | 0.9174 | 0.8223 | 0.6587 |
|  | LightGBM | RMSE | 0.2421 | 0.3763 | 0.5546 | 0.7690 |
|  |  | MAE | 0.1599 | 0.2477 | 0.3708 | 0.5281 |
|  |  | MAPE | 2.3568 | 3.9546 | 5.1076 | 7.1759 |
|  |  | R | 0.9654 | 0.9164 | 0.8185 | 0.6509 |
|  | LSTM | RMSE | 0.2332 | 0.3308 | 0.4615 | 0.7055 |
|  |  | MAE | 0.1623 | 0.2053 | 0.3155 | 0.4699 |
|  |  | MAPE | 1.9646 | 2.0817 | 2.5948 | 2.9523 |
|  |  | R | 0.9740 | 0.9468 | 0.8964 | 0.8032 |
|  | TCN | RMSE | 0.2230 | 0.2948 | 0.4423 | 0.6217 |
|  |  | MAE | 0.1322 | 0.1987 | 0.3255 | 0.4392 |
|  |  | MAPE | 1.3568 | 1.5604 | 1.9076 | 2.7701 |
|  |  | R | 0.9822 | 0.9620 | 0.9063 | 0.8268 |
|  | MWNet | RMSE | 0.2250 | 0.3095 | 0.4565 | 0.6282 |
|  |  | MAE | 0.1397 | 0.2099 | 0.2987 | 0.4440 |
|  |  | MAPE | 1.6769 | 1.6462 | 2.2063 | 2.7710 |
|  |  | R | 0.9818 | 0.9562 | 0.9015 | 0.8246 |
|  | AFE-TFNet | RMSE | 0.1821 | 0.2341 | 0.3156 | 0.4409 |
|  |  | MAE | 0.1231 | 0.1538 | 0.2054 | 0.2860 |
|  |  | MAPE | 1.1901 | 1.3927 | 1.9257 | 2.6173 |
|  |  | R | 0.9842 | 0.9740 | 0.9524 | 0.9078 |

Table 6 Performance metrics of all models across dataset B.

| Station | Model | Metrics | Prediction time | | | |
|---|---|---|---|---|---|---|
|  |  |  | 1h | 3h | 6h | 12h |
| B | NaiveDrift | RMSE | 0.2822 | 0.3958 | 0.5210 | 0.6815 |
|  |  | MAE | 0.1937 | 0.2506 | 0.3086 | 0.4135 |
|  |  | MAPE | 1.9173 | 2.9136 | 3.1450 | 8.7117 |
|  |  | R | 0.9729 | 0.9075 | 0.8618 | 0.6892 |
|  | XGBoost | RMSE | 0.2731 | 0.3596 | 0.4625 | 0.6098 |
|  |  | MAE | 0.1518 | 0.1933 | 0.2960 | 0.3926 |
|  |  | MAPE | 1.1679 | 2.6315 | 4.6055 | 8.3605 |
|  |  | R | 0.9500 | 0.9097 | 0.8238 | 0.6516 |
|  | CatBoost | RMSE | 0.2950 | 0.3321 | 0.4286 | 0.6756 |
|  |  | MAE | 0.1640 | 0.2115 | 0.3087 | 0.4182 |
|  |  | MAPE | 1.8425 | 2.9815 | 3.5307 | 7.5638 |
|  |  | R | 0.9635 | 0.9278 | 0.8552 | 0.6967 |
|  | LightGBM | RMSE | 0.2542 | 0.3242 | 0.4242 | 0.6709 |
|  |  | MAE | 0.1533 | 0.2016 | 0.3087 | 0.4075 |
|  |  | MAPE | 1.0288 | 2.0310 | 3.6005 | 7.6310 |
|  |  | R | 0.9681 | 0.9326 | 0.8591 | 0.7027 |
|  | LSTM | RMSE | 0.2427 | 0.3744 | 0.4747 | 0.7071 |
|  |  | MAE | 0.1292 | 0.2054 | 0.2745 | 0.4156 |
|  |  | MAPE | 0.7154 | 1.0588 | 1.3559 | 1.8814 |
|  |  | R | 0.9781 | 0.9504 | 0.9127 | 0.8150 |
|  | TCN | RMSE | 0.1789 | 0.2764 | 0.4030 | 0.6119 |
|  |  | MAE | 0.1165 | 0.1764 | 0.2605 | 0.3894 |
|  |  | MAPE | 0.6475 | 0.9241 | 1.2660 | 1.8059 |
|  |  | R | 0.9866 | 0.9630 | 0.9338 | 0.8437 |
|  | MWNet | RMSE | 0.2365 | 0.2532 | 0.3937 | 0.6223 |
|  |  | MAE | 0.1435 | 0.1638 | 0.2461 | 0.4017 |
|  |  | MAPE | 0.6570 | 0.8457 | 1.2609 | 1.6804 |
|  |  | R | 0.9806 | 0.9610 | 0.9286 | 0.8383 |
|  | AFE-TFNet | RMSE | 0.1693 | 0.2071 | 0.2735 | 0.3930 |
|  |  | MAE | 0.1113 | 0.1328 | 0.1729 | 0.2420 |
|  |  | MAPE | 0.6025 | 0.7396 | 0.9597 | 1.3027 |
|  |  | R | 0.9873 | 0.9803 | 0.9655 | 0.9290 |

Table 7 Performance metrics of all models across dataset C.

| Station | Model | Metrics | Prediction time |
|---|---|---|---|

|  |  |  | 1h | 3h | 6h | 12h |
|---|---|---|---|---|---|---|
| C | NaiveDrift | RMSE | 0.3010 | 0.4649 | 0.7444 | 0.9343 |
|  |  | MAE | 0.3925 | 0.4159 | 0.7686 | 0.9309 |
|  |  | MAPE | 5.1837 | 7.7616 | 9.0203 | 13.0889 |
|  |  | R | 0.9547 | 0.8793 | 0.7348 | 0.5034 |
|  | XGBoost | RMSE | 0.3095 | 0.4746 | 0.6481 | 0.7295 |
|  |  | MAE | 0.3755 | 0.4199 | 0.6702 | 0.7263 |
|  |  | MAPE | 3.3831 | 4.1295 | 5.4131 | 7.4262 |
|  |  | R | 0.9468 | 0.8647 | 0.7266 | 0.5177 |
|  | CatBoost | RMSE | 0.2996 | 0.3593 | 0.6322 | 0.7104 |
|  |  | MAE | 0.3709 | 0.4115 | 0.6582 | 0.7127 |
|  |  | MAPE | 3.0592 | 4.4875 | 5.3906 | 7.1353 |
|  |  | R | 0.9560 | 0.8873 | 0.7605 | 0.5718 |
|  | LightGBM | RMSE | 0.2995 | 0.3608 | 0.6350 | 0.7128 |
|  |  | MAE | 0.3410 | 0.4124 | 0.5600 | 0.6147 |
|  |  | MAPE | 3.0634 | 3.5398 | 5.5257 | 7.3038 |
|  |  | R | 0.9560 | 0.8853 | 0.7548 | 0.5653 |
|  | LSTM | RMSE | 0.3657 | 0.4273 | 0.6359 | 0.7439 |
|  |  | MAE | 0.3167 | 0.3492 | 0.5364 | 0.6039 |
|  |  | MAPE | 3.2431 | 3.6411 | 5.0651 | 5.1319 |
|  |  | R | 0.9715 | 0.9255 | 0.8384 | 0.6993 |
|  | TCN | RMSE | 0.1966 | 0.3347 | 0.5128 | 0.6631 |
|  |  | MAE | 0.1442 | 0.2399 | 0.3512 | 0.4588 |
|  |  | MAPE | 1.7593 | 2.7229 | 3.5901 | 4.4788 |
|  |  | R | 0.9770 | 0.9309 | 0.8378 | 0.7040 |
|  | MWNet | RMSE | 0.1989 | 0.3357 | 0.5269 | 0.6718 |
|  |  | MAE | 0.1466 | 0.2377 | 0.3515 | 0.4605 |
|  |  | MAPE | 1.7298 | 2.7955 | 3.3966 | 3.8829 |
|  |  | R | 0.9769 | 0.9318 | 0.8452 | 0.6908 |
|  | AFE-TFNet | RMSE | 0.1962 | 0.2874 | 0.3923 | 0.5405 |
|  |  | MAE | 0.1435 | 0.2008 | 0.2639 | 0.3623 |
|  |  | MAPE | 0.8104 | 1.0519 | 1.3094 | 1.8037 |
|  |  | R | 0.9793 | 0.9630 | 0.9306 | 0.8735 |

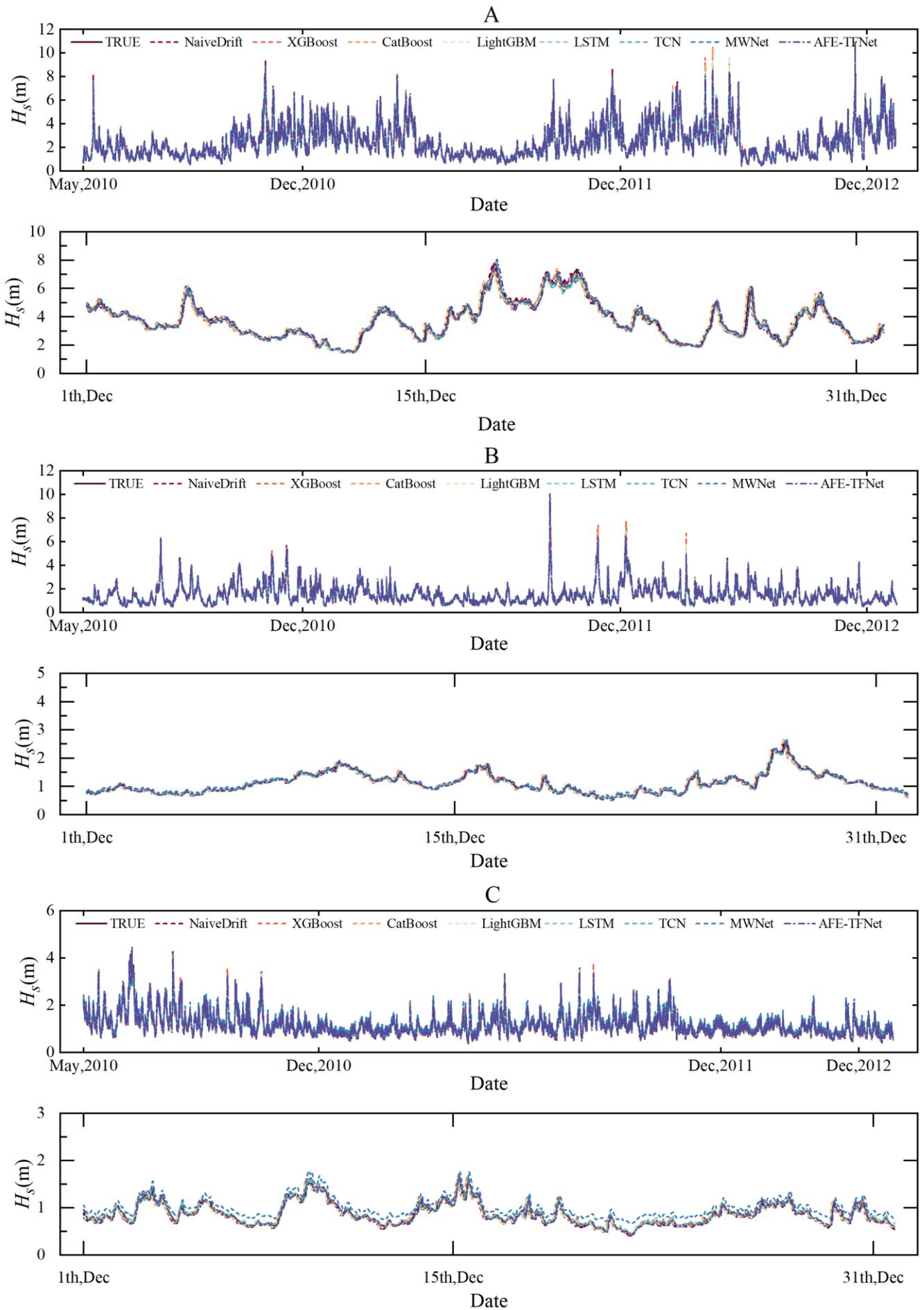

Fig. 7. Comparisons of observation values and the 1-hour prediction of the models.

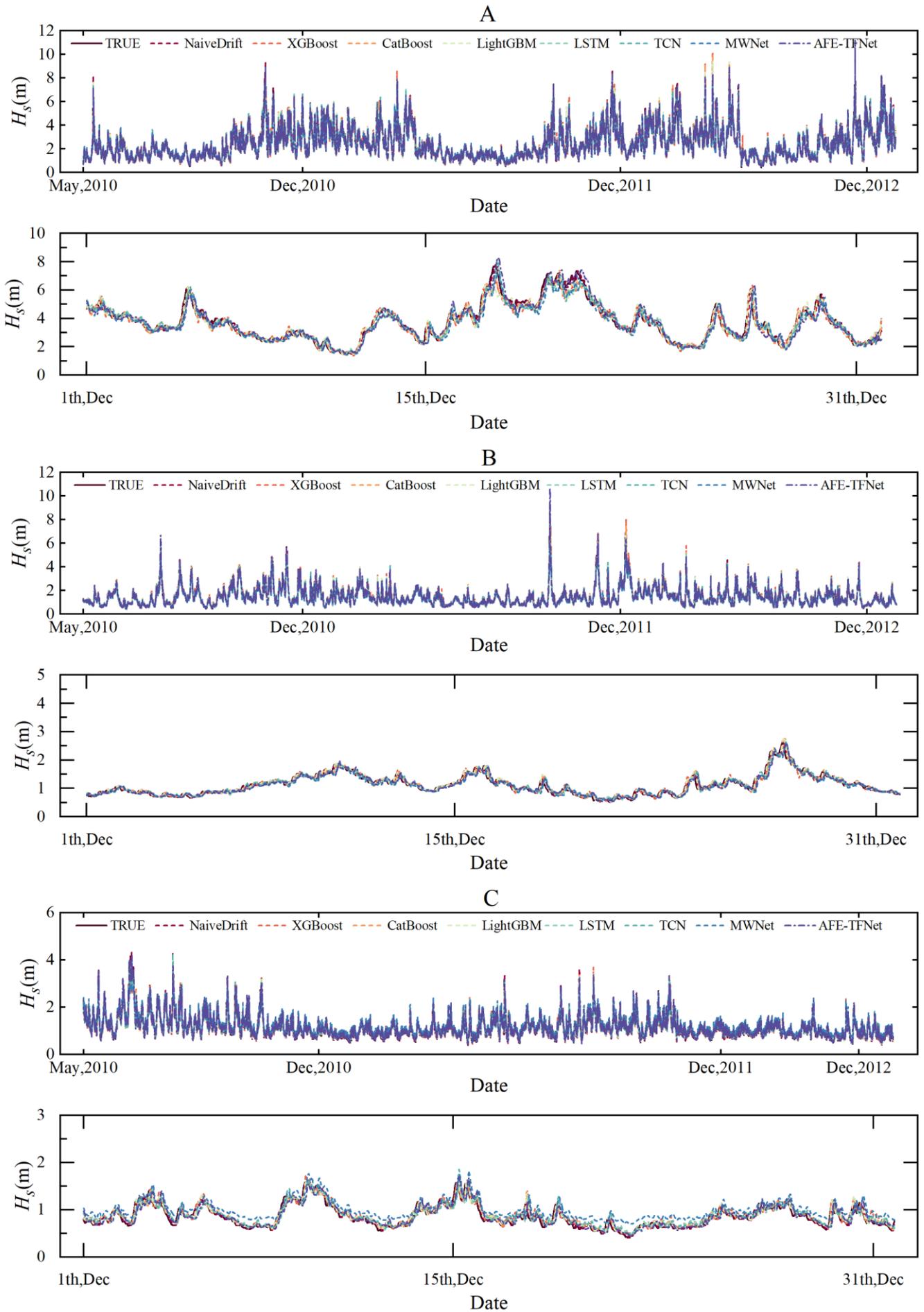

Fig. 8. Comparisons of observation values and the 3-hour prediction of the models.

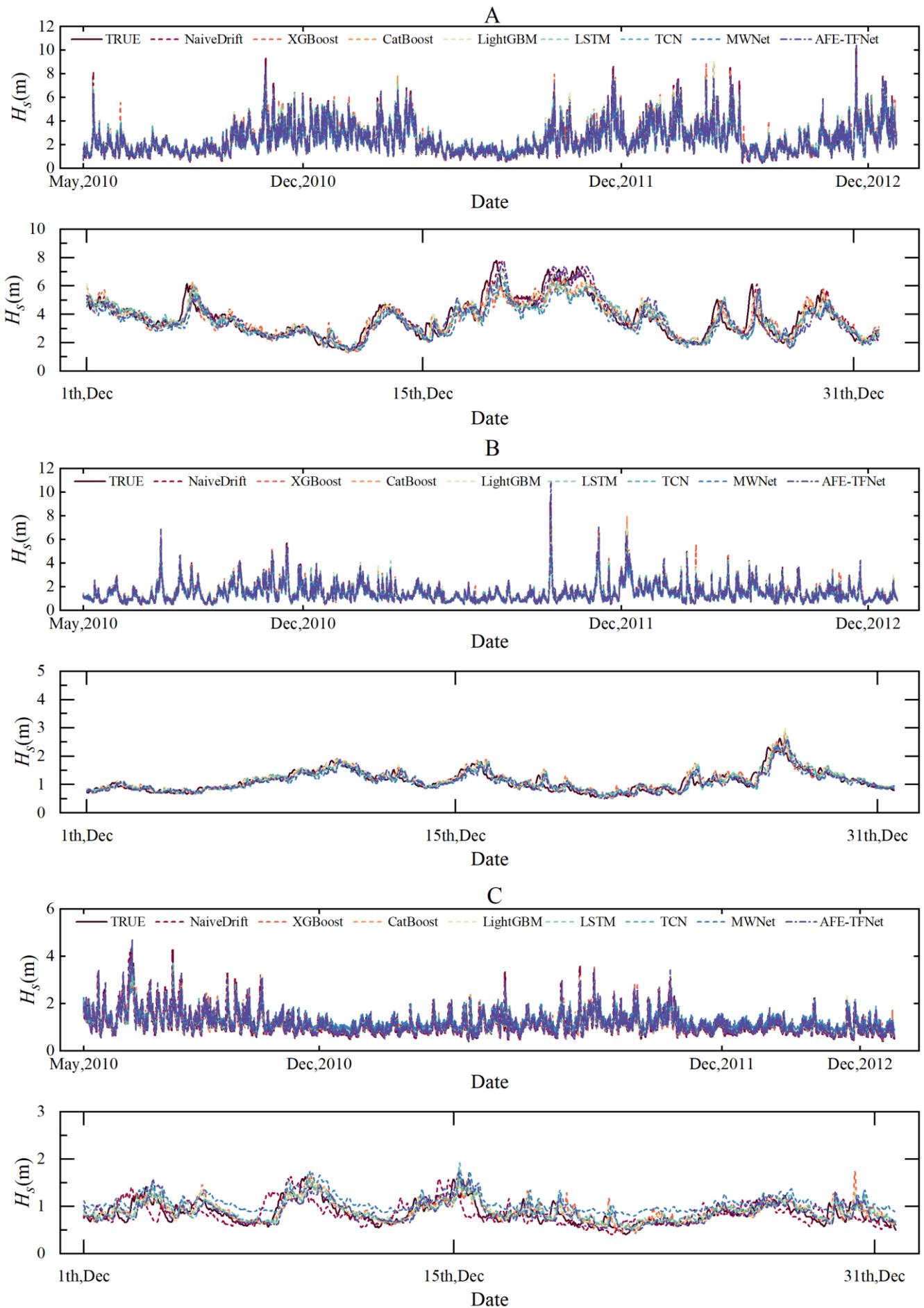

Fig. 9. Comparisons of observation values and the 6-hour prediction of the models.

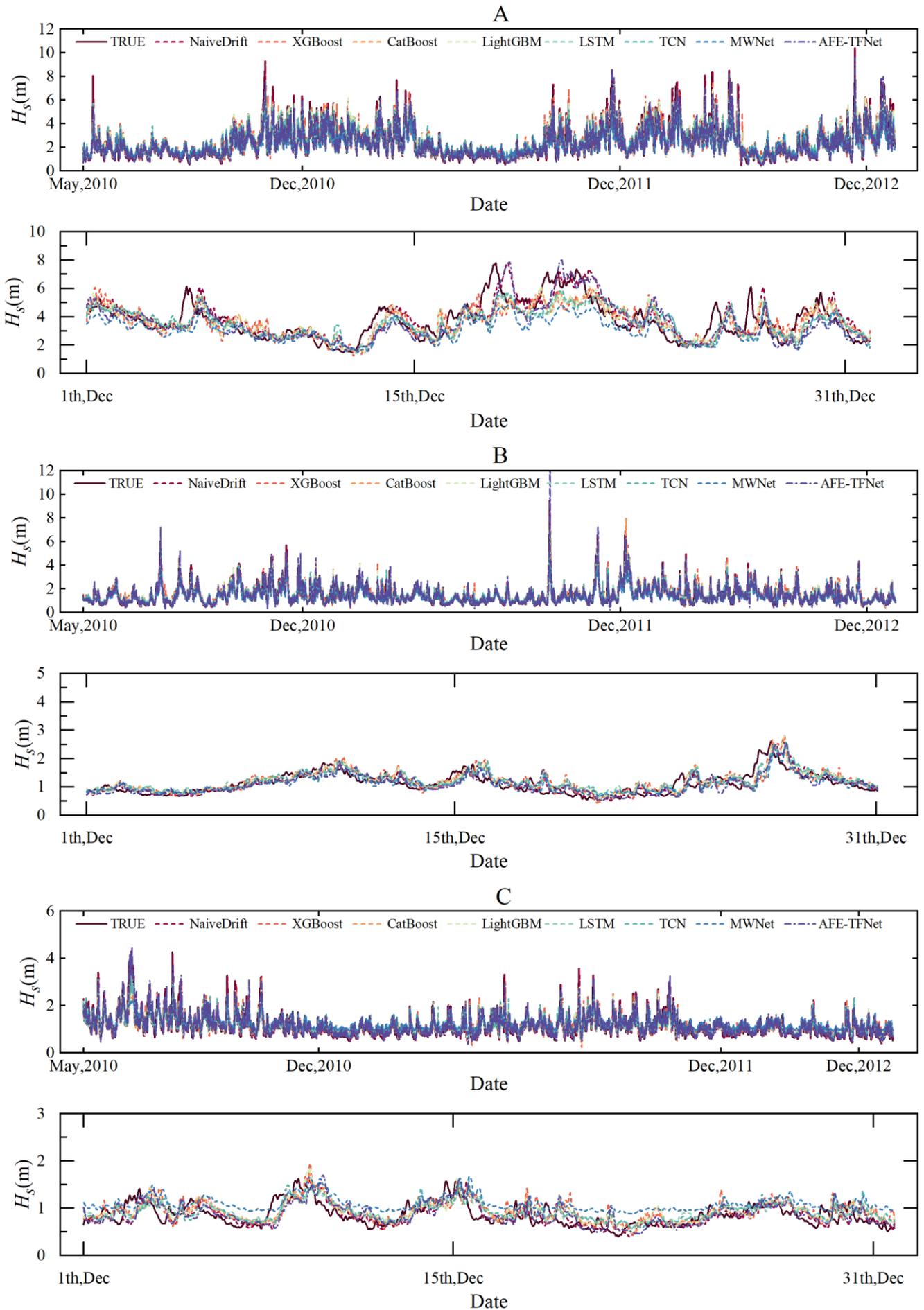

Fig. 10. Comparisons of observation values and the 12-hour prediction of the models.

This study further evaluates the performance of AFE-TFNet through scatter plots and box plots, as shown in Fig. 11 and

Fig. 12. These visualizations provide a deeper insight into the model's error distribution and stability across different prediction scenarios. The scatter points in Fig. 11 represent the predicted wave heights (y-axis) plotted against the observed wave heights (x-axis). As the prediction time increases, the scatter points gradually deviate from the diagonal line. This deviation is most pronounced in the 12 h prediction. This indicates a general decrease in prediction accuracy over time. AFE-TFNet shows better alignment with observed trends in the 6 h and 12 h predictions, with less scatter, indicating greater stability in long-term forecasts. From the distribution of scatter points, those below the diagonal represent underestimation, while those above represent overestimation. In the short-term predictions of 1 h and 3 h, all models show relatively small errors and stable performance. For the 6 h and 12 h predictions, when wave heights are low, other models tend to exhibit larger overestimations or underestimations, whereas AFE-TFNet remains more stable. Notably, when the values observed are high, AFE-TFNet tends to overestimate, while other models tend to underestimate. This behavior may be attributed to AFE-TFNet's heightened sensitivity to local fluctuations, which results in a stronger response to complex cyclical variations. In contrast, other models rely more on smoother trends in time-series data, which struggle to capture these rare wave heights.

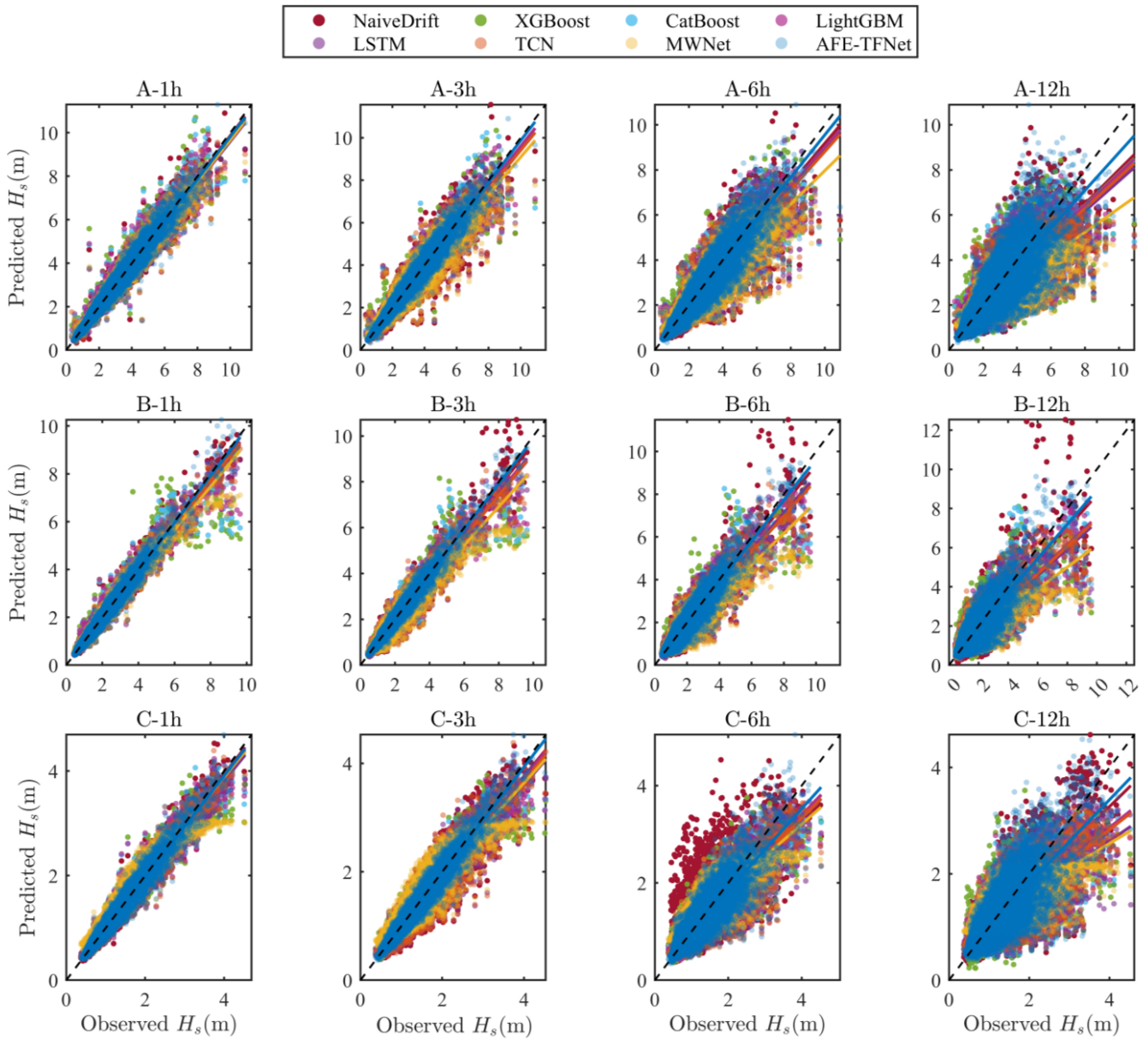

Fig. 11. Scatter plot comparing observed and predicted $H_s$.

To further evaluate the performance in capturing these rare wave heights, Fig. 12 shows box plots comparing the predictions with the observed values. Each boxplot shows the distribution of $H_s$ values for the corresponding model and for the true observations across the test set. The outliers in the box plots represent extreme value predictions. Comparative analysis of various models reveals that AFE-TFNet demonstrates superior consistency with observed values in both short-term and long-term predictions, except for the Naive Drift model. This exception can be attributed to the Naive Drift model's "lazy" preservation of temporal inertia or simple drift in the original distribution, which may result in similar distribution patterns and outlier positions to the actual values.

Overall, compared to other models, AFE-TFNet performs exceptionally well in long-term forecasts and in predicting rare wave heights. This indicates AFE-TFNet's dynamic integration of time- and frequency-domain information through the DHSEW mechanism. The dual-domain integration enables AFE-TFNet to capture various signal patterns.

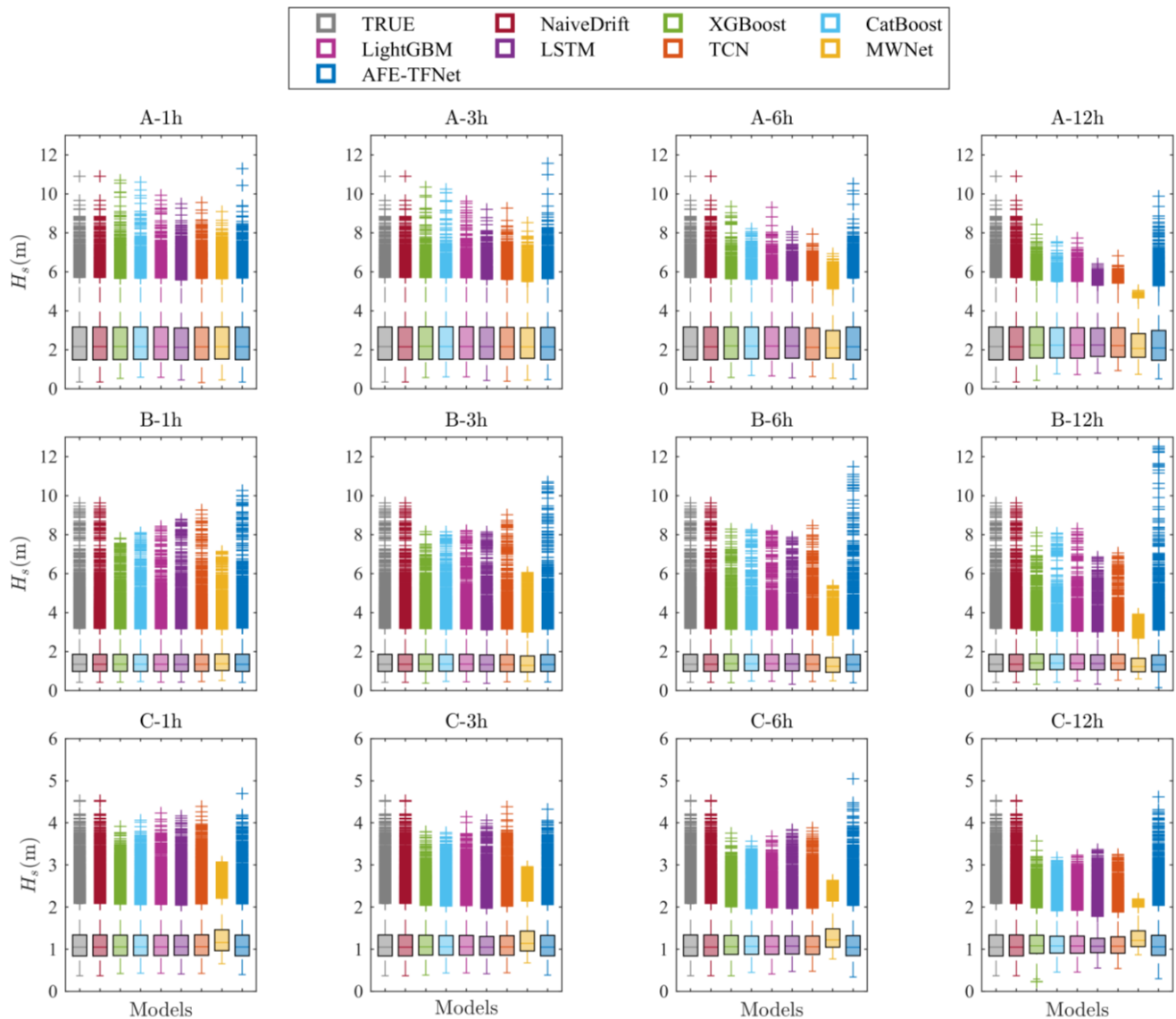

Fig. 12. Boxplots of observed and predicted $H_s$.

**4.4 Seasonal performance analysis**

To further assess the performance of AFE-TFNet under varying environmental conditions, we conducted seasonal forecasting experiments for spring, summer, autumn, and winter. The seasonal performance metrics (*RMSE*, *MAE*, *MAPE*, and *R*) are summarized in Table 8, 9, and 10. Analysis reveals that AFE-TFNet consistently achieves superior performance across all three datasets and four seasons, maintaining the lowest *RMSE*, *MAE*, and *MAPE* values, while exhibiting the highest *R* values compared to all baseline models. This consistent outperformance validates AFE-TFNet's robust capacity for significant wave height forecasting, independent of seasonal variations.

While AFE-TFNet demonstrates consistent superiority over baseline models across all seasons, the magnitude of its performance improvement exhibits seasonal variation. For Dataset A, AFE-TFNet achieves an average *RMSE* reduction of approximately 3.0% compared to the best-performing baseline during spring and summer, with this improvement increasing to approximately 5.0% during autumn and winter. Similarly, for Dataset C, the *RMSE* reduction increases from about 4.5% in spring and summer to approximately 7.5% in autumn and winter. This pattern indicates that AFE-TFNet's performance advantage, particularly in terms of *RMSE*, is more pronounced during autumn and winter months for both Datasets A and C. These findings suggest that AFE-TFNet's sophisticated feature extraction and time-frequency analysis capabilities are particularly effective in capturing the more complex and variable patterns characteristic of autumn and winter seasons.

Table 8 Performance metrics of all models across various dataset A.

| Station | Season | Model | RMSE | MAE | MAPE | R |
|---|---|---|---|---|---|---|
| A | Spring | NaiveDrift | 0.1598 | 0.1117 | 6.3110 | 0.9758 |
| | | XGBoost | 0.1527 | 0.1204 | 6.7437 | 0.9777 |
| | | CatBoost | 0.1500 | 0.1345 | 6.4244 | 0.9713 |
| | | LightGBM | 0.1501 | 0.1458 | 6.5374 | 0.9713 |
| | | LSTM | 0.1495 | 0.1165 | 6.3145 | 0.9791 |
| | | TCN | 0.1477 | 0.1152 | 6.1633 | 0.9793 |
| | | MWNet | 0.1580 | 0.1145 | 6.9103 | 0.9777 |
| | | AFE-TFNet | 0.1437 | 0.1067 | 6.0550 | 0.9884 |
| | Summer | NaiveDrift | 0.2286 | 0.1519 | 6.3937 | 0.9840 |
| | | XGBoost | 0.2277 | 0.1590 | 6.8001 | 0.9841 |
| | | CatBoost | 0.2133 | 0.1520 | 6.5369 | 0.9862 |
| | | LightGBM | 0.2070 | 0.1522 | 6.6068 | 0.9870 |
| | | LSTM | 0.2178 | 0.1427 | 6.8847 | 0.9862 |
| | | TCN | 0.2149 | 0.1420 | 6.9208 | 0.9860 |
| | | MWNet | 0.2250 | 0.1540 | 7.0501 | 0.9849 |
| | | AFE-TFNet | 0.2033 | 0.1418 | 6.0191 | 0.9880 |
| | Autumn | NaiveDrift | 0.2822 | 0.2066 | 6.6724 | 0.9609 |
| | | XGBoost | 0.2611 | 0.1820 | 6.9380 | 0.9662 |
| | | CatBoost | 0.2634 | 0.1716 | 6.5495 | 0.9606 |
| | | LightGBM | 0.2674 | 0.1739 | 6.6280 | 0.9697 |
| | | LSTM | 0.2638 | 0.1900 | 6.0919 | 0.9660 |
| | | TCN | 0.2626 | 0.1892 | 6.0678 | 0.9659 |
| | | MWNet | 0.2694 | 0.1977 | 6.3064 | 0.9645 |
| | | AFE-TFNet | 0.2561 | 0.1642 | 6.0434 | 0.9750 |
| | Winter | NaiveDrift | 0.2128 | 0.1388 | 6.0577 | 0.9756 |
| | | XGBoost | 0.2004 | 0.1355 | 6.5246 | 0.9772 |
| | | CatBoost | 0.1942 | 0.1362 | 6.1183 | 0.9792 |
| | | LightGBM | 0.1952 | 0.1394 | 6.2891 | 0.9791 |
| | | LSTM | 0.2032 | 0.1330 | 5.8462 | 0.9852 |
| | | TCN | 0.1932 | 0.1282 | 5.7166 | 0.9851 |
| | | MWNet | 0.2060 | 0.1383 | 6.5876 | 0.9838 |
| | | AFE-TFNet | 0.1895 | 0.1230 | 5.6766 | 0.9873 |

Table 9 Performance metrics of all models across various dataset B.

| Station | Season | Model | RMSE | MAE | MAPE | R |
|---|---|---|---|---|---|---|
| B | Spring | NaiveDrift | 0.3342 | 0.2245 | 7.2915 | 0.8616 |
| | | XGBoost | 0.3285 | 0.2199 | 7.0496 | 0.8659 |
| | | CatBoost | 0.1334 | 0.1218 | 6.0330 | 0.9641 |
| | | LightGBM | 0.1383 | 0.1194 | 6.9655 | 0.9657 |
| | | LSTM | 0.1328 | 0.0949 | 5.5574 | 0.9686 |
| | | TCN | 0.1326 | 0.0939 | 5.7131 | 0.9753 |
| | | MWNet | 0.1417 | 0.1023 | 6.4123 | 0.9759 |
| | | AFE-TFNet | 0.1323 | 0.0928 | 5.5261 | 0.9783 |
| | Summer | NaiveDrift | 0.1211 | 0.1588 | 8.3729 | 0.8736 |
| | | XGBoost | 0.1180 | 0.1570 | 8.2396 | 0.8764 |
| | | CatBoost | 0.1008 | 0.0677 | 6.1969 | 0.9735 |
| | | LightGBM | 0.1089 | 0.0670 | 6.2096 | 0.9744 |
| | | LSTM | 0.0858 | 0.0624 | 5.2720 | 0.9804 |

|  | | Model | RMSE | MAE | MAPE | R |
|---|---|---|---|---|---|---|
| | | TCN | 0.0860 | 0.0630 | 5.2432 | 0.9804 |
| | | MWNet | 0.0999 | 0.0795 | 7.3020 | 0.9786 |
| | | AFE-TFNet | 0.0835 | 0.0540 | 5.1247 | 0.9843 |
| | Autumn | NaiveDrift | 0.2171 | 0.1209 | 8.6473 | 0.8681 |
| | | XGBoost | 0.2945 | 0.1163 | 8.4895 | 0.8824 |
| | | CatBoost | 0.1259 | 0.0806 | 6.2975 | 0.9781 |
| | | LightGBM | 0.1246 | 0.0811 | 6.4315 | 0.9774 |
| | | LSTM | 0.1260 | 0.0789 | 5.8922 | 0.9889 |
| | | TCN | 0.1326 | 0.0793 | 5.8674 | 0.9883 |
| | | MWNet | 0.1880 | 0.0976 | 7.5206 | 0.9830 |
| | | AFE-TFNet | 0.1219 | 0.0775 | 5.8081 | 0.9896 |
| | Winter | NaiveDrift | 1.1181 | 0.8497 | 8.3333 | 0.8264 |
| | | XGBoost | 1.1265 | 0.8519 | 8.2762 | 0.8279 |
| | | CatBoost | 0.1847 | 0.1487 | 6.1014 | 0.9820 |
| | | LightGBM | 0.1892 | 0.1452 | 6.9375 | 0.9820 |
| | | LSTM | 0.1721 | 0.1127 | 5.4591 | 0.9867 |
| | | TCN | 0.1754 | 0.1140 | 5.5614 | 0.9863 |
| | | MWNet | 0.1857 | 0.1224 | 6.2569 | 0.9851 |
| | | AFE-TFNet | 0.1693 | 0.1021 | 5.4326 | 0.9870 |

Table 10 Performance metrics of all models across various dataset C.

| Station | Season | Model | Metrics | | | |
|---|---|---|---|---|---|---|
| | | | RMSE | MAE | MAPE | R |
| C | Spring | NaiveDrift | 0.2029 | 0.4488 | 15.6146 | 0.8668 |
| | | XGBoost | 0.2002 | 0.4467 | 15.3997 | 0.8660 |
| | | CatBoost | 0.0976 | 0.0755 | 7.1767 | 0.9675 |
| | | LightGBM | 0.0986 | 0.0784 | 7.0634 | 0.9694 |
| | | LSTM | 0.0963 | 0.0721 | 6.1724 | 0.9760 |
| | | TCN | 0.0968 | 0.0727 | 6.1765 | 0.9755 |
| | | MWNet | 0.1682 | 0.1430 | 13.4492 | 0.9673 |
| | | AFE-TFNet | 0.0923 | 0.0619 | 6.1507 | 0.9783 |
| | Summer | NaiveDrift | 0.5319 | 0.3921 | 14.4754 | 0.8761 |
| | | XGBoost | 0.5312 | 0.3916 | 14.4350 | 0.8770 |
| | | CatBoost | 0.1281 | 0.0897 | 5.7339 | 0.9727 |
| | | LightGBM | 0.1362 | 0.0981 | 5.7136 | 0.9733 |
| | | LSTM | 0.0922 | 0.0674 | 5.6372 | 0.9753 |
| | | TCN | 0.0933 | 0.0678 | 5.6267 | 0.9748 |
| | | MWNet | 0.1622 | 0.1388 | 12.9599 | 0.9706 |
| | | AFE-TFNet | 0.0817 | 0.0572 | 5.5150 | 0.9757 |
| | Autumn | NaiveDrift | 0.1733 | 0.2225 | 13.0683 | 0.8653 |
| | | XGBoost | 0.1734 | 0.2227 | 13.0769 | 0.8695 |
| | | CatBoost | 0.0738 | 0.1230 | 6.1024 | 0.9369 |
| | | LightGBM | 0.0779 | 0.1276 | 6.9464 | 0.9366 |
| | | LSTM | 0.0693 | 0.0540 | 5.5163 | 0.9427 |
| | | TCN | 0.0680 | 0.0525 | 5.3155 | 0.9453 |
| | | MWNet | 0.1434 | 0.1273 | 13.9381 | 0.9318 |
| | | AFE-TFNet | 0.0576 | 0.0513 | 5.2857 | 0.9591 |
| | Winter | NaiveDrift | 0.2104 | 0.3011 | 14.1192 | 0.8705 |
| | | XGBoost | 0.2095 | 0.3004 | 14.0177 | 0.8713 |
| | | CatBoost | 0.0969 | 0.0994 | 6.8249 | 0.9632 |
| | | LightGBM | 0.0910 | 0.0814 | 6.6361 | 0.9650 |
| | | LSTM | 0.0927 | 0.0626 | 6.7023 | 0.9688 |
| | | TCN | 0.0920 | 0.0611 | 6.5286 | 0.9688 |
| | | MWNet | 0.1658 | 0.1474 | 17.8443 | 0.9630 |
| | | AFE-TFNet | 0.0814 | 0.0578 | 6.3939 | 0.9798 |

Figs. 13-15 display comparative curves between predicted and actual wave heights across datasets A, B, and C for all four seasons. During spring and summer, the prediction curves for all models closely follow the true wave height curves, indicating relatively smooth and predictable wave behavior. In contrast, the autumn and winter panels reveal substantially more dynamic and energetic wave behaviors. These seasons are characterized by elevated wave heights and more pronounced

fluctuations, manifested through frequent and sharp peaks and troughs, indicating increased wave intensity and variability. AFE-TFNet exhibits particularly robust performance in Datasets A and C during autumn and winter periods, where its prediction curves maintain stronger correspondence with actual wave heights, especially during high-intensity events and periods of significant variability. This contrasts with baseline models, which frequently exhibit either overestimation or underestimation of these dynamic variations. The superior performance of AFE-TFNet can be attributed to its advanced feature extraction and time-frequency analysis capabilities, specifically its DHSEW mechanisms. These components demonstrate effectiveness in processing the complex, variable, and aperiodic wave dynamics characteristic of autumn and winter conditions.

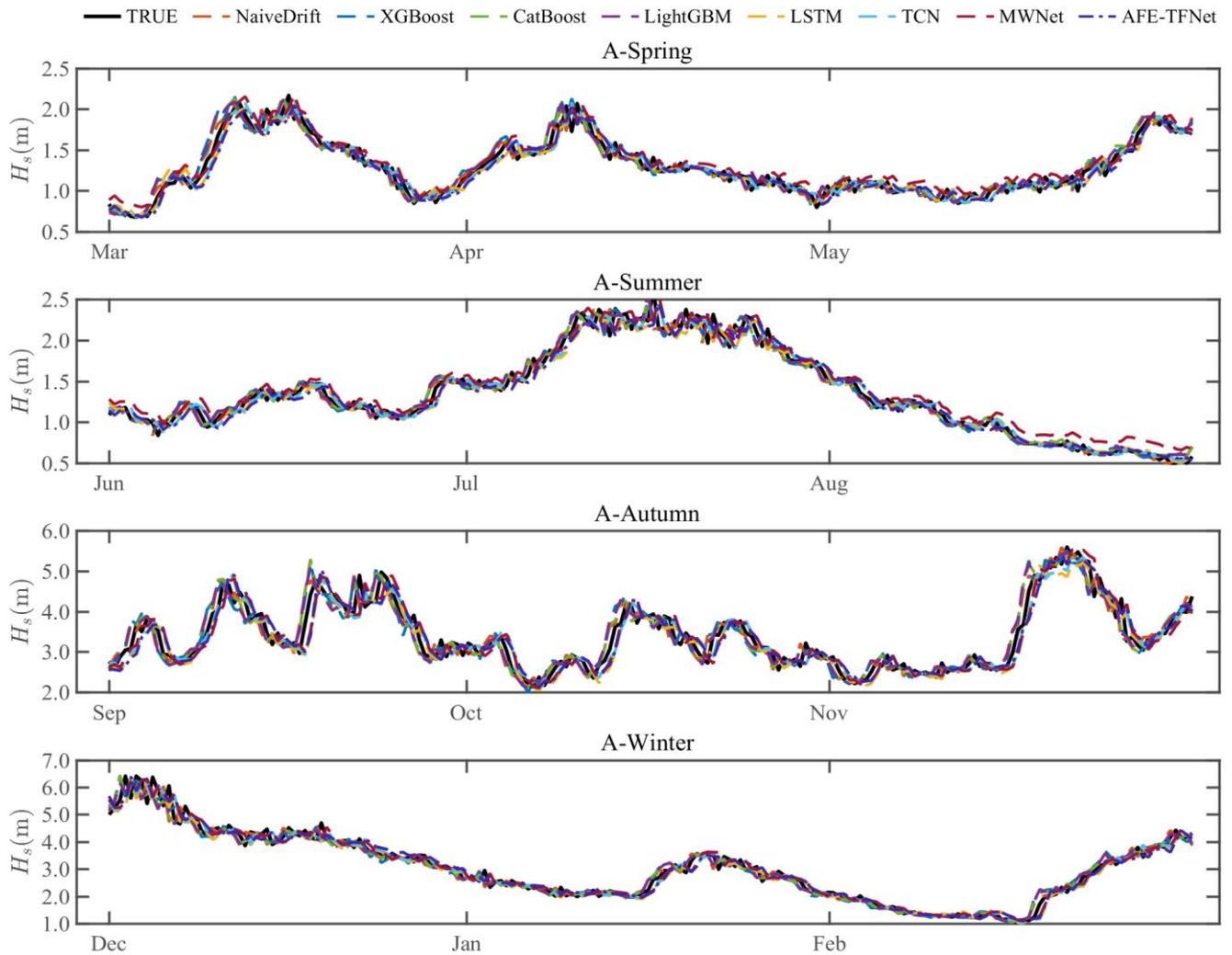

Fig. 13. Seasonal wave height prediction comparison for models on dataset A.

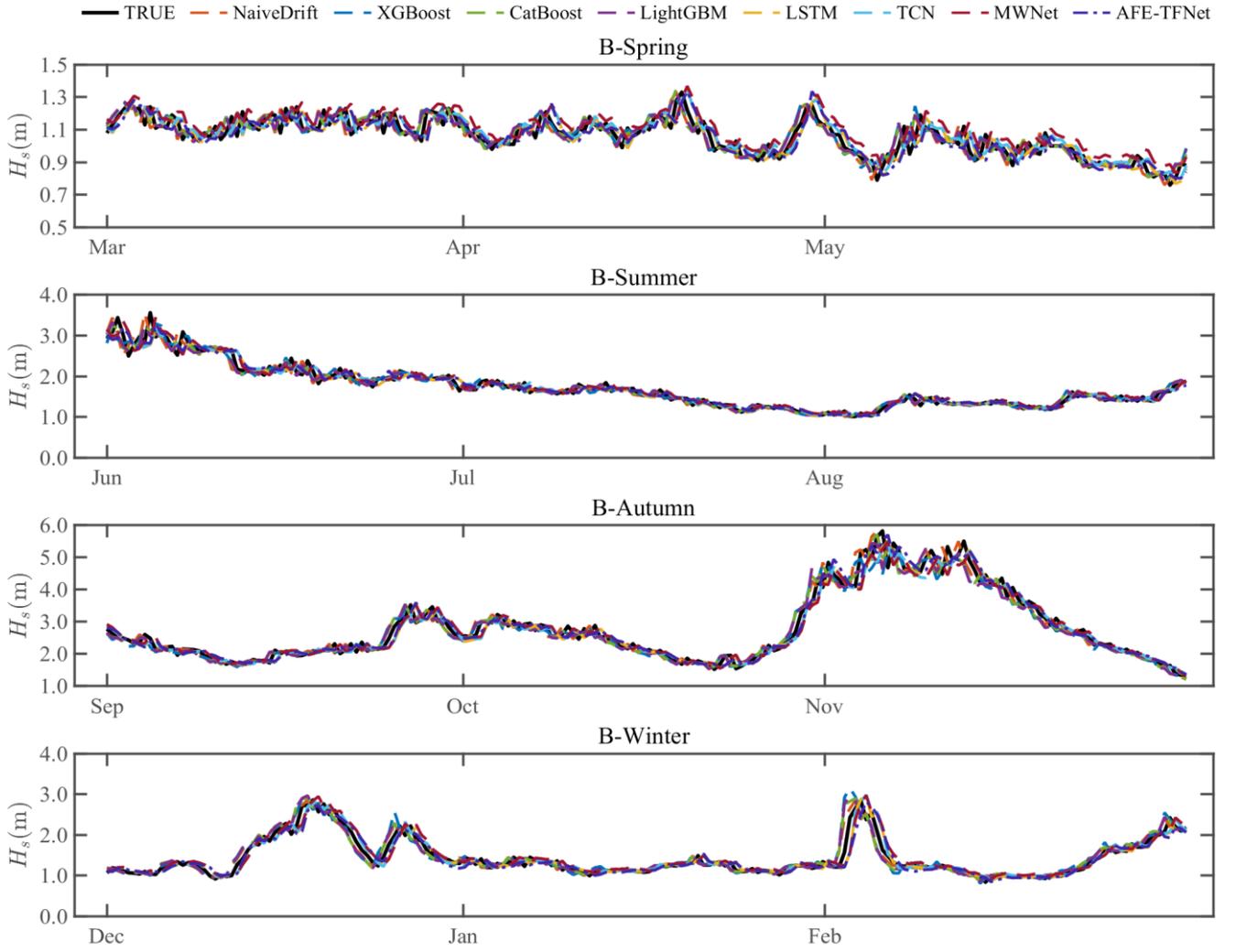

Fig. 14. Seasonal wave height prediction comparison for models on dataset B.

**4.5 Interval prediction experiment**

To extend our evaluation beyond point forecasting, we conducted interval forecasting experiments using the Bootstrap method, a non-parametric resampling technique that estimates prediction uncertainty by generating multiple simulated forecast trajectories. We applied this method to both AFE-TFNet and all baseline models, maintaining the same training procedure described in Section 4.3.

**4.5.1 Evaluation metrics**

The quality of prediction intervals was assessed using two standard metrics:

Prediction Interval Coverage Probability (PICP): PICP measures the percentage of observed values that fall within the predicted prediction intervals. A higher PICP value indicates better coverage. The metric is defined as:

$$PICP = \frac{1}{n}\sum_{i=1}^{n} I\left(y_i \in [L_i, U_i]\right) \tag{19}$$

where $n$ is the total number of samples, $y_i$ is the $i$-th observed value, and $[L_i, U_i]$ is the lower and upper bounds of the $i$-th prediction interval.

Prediction Interval Normalized Average Width (PINAW): PINAW measures the average width of the prediction intervals, normalized by the range of the observed data. A lower PINAW value indicates narrower and more informative intervals. The metric is defined as:

$$PINAW = \frac{1}{n}\sum_{i=1}^{n} \frac{U_i - L_i}{\max(y) - \min(y)} \tag{20}$$

where $n$ is the total number of samples, $L_i$ and $U_i$ are the lower and upper bounds of the $i$-th prediction interval, and $\max(y)$ and $\min(y)$ are the maximum and minimum values of the observed data.

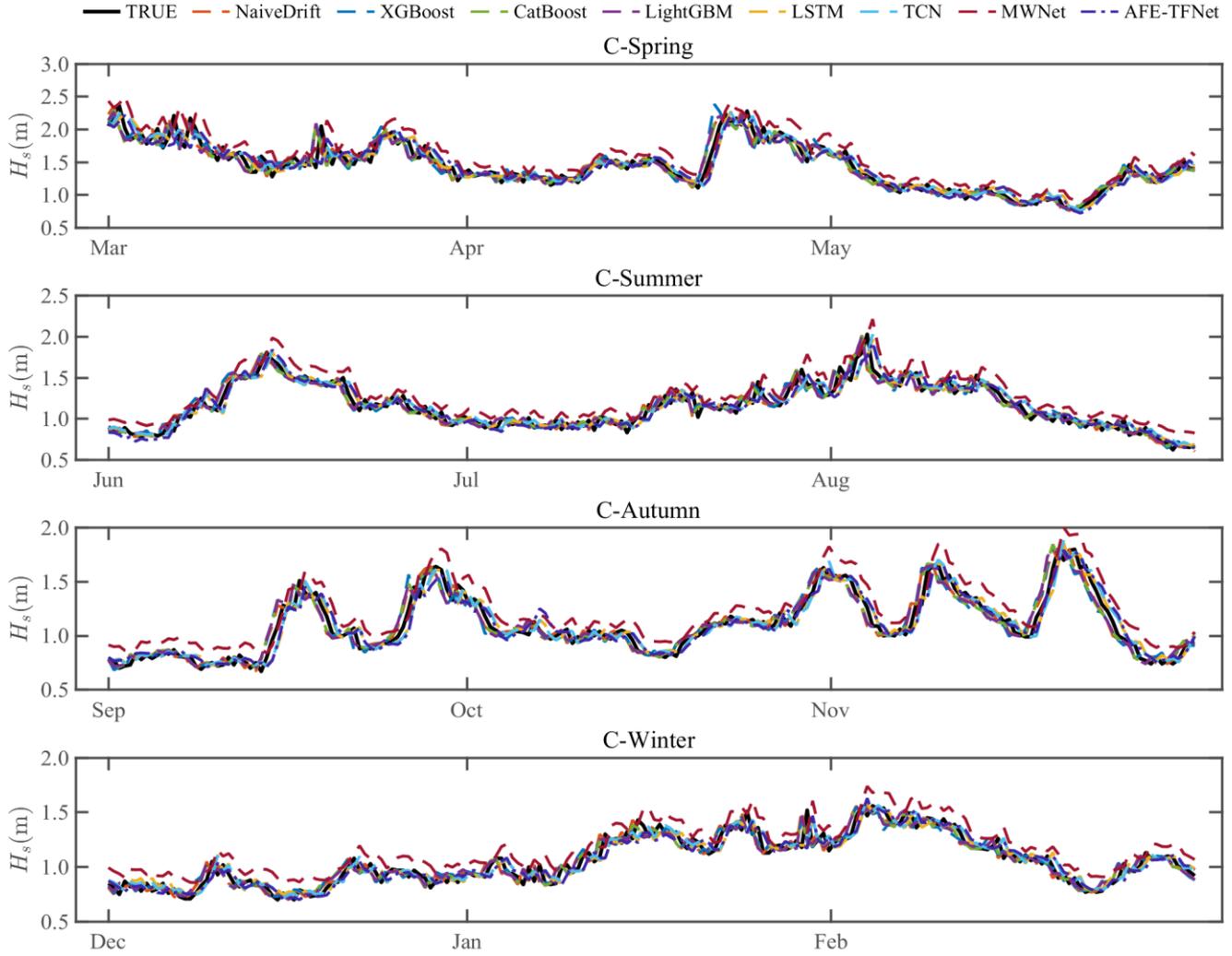

Fig. 15. Seasonal wave height prediction comparison for models on dataset C.

### 4.5.2. Results and discussion

Table 11 presents the PICP and PINAW values for AFE-TFNet and baseline models across the three datasets at 85%, 90%, and 95% confidence levels. Focusing on the standard 90% confidence level, AFE-TFNet consistently achieves PICP values near or exceeding 90% across all datasets, demonstrating well-calibrated prediction intervals. While LSTM and TCN show comparable PICP values, traditional machine learning models and NaiveDrift exhibit slightly lower coverage, particularly for Dataset C. Regarding PINAW, AFE-TFNet consistently achieves the lowest values across all datasets at 90% confidence, indicating narrower, more informative prediction intervals while maintaining robust coverage probability. For instance, on Dataset B at 90% confidence, AFE-TFNet achieves a PINAW of 1.3284, which is notably narrower than LSTM's 1.3970, TCN's 1.4008, and MWNet's 1.3312, while still maintaining a PICP exceeding 90%. Traditional machine learning models and NaiveDrift consistently show the highest PINAW values, indicating less efficient uncertainty estimation.

Table 11 Performance metrics of all models.

| Station | Model | Confidence level | | | | | |
| --- | --- | --- | --- | --- | --- | --- | --- |
| | | 85% | | 90% | | 95% | |
| | | PINAW | PICP | PINAW | PICP | PINAW | PICP |
| A | NaiveDrift | 3.1612 | 0.7532 | 3.6102 | 0.8301 | 4.3581 | 0.9125 |
| | XGBoost | 3.1662 | 0.7542 | 3.6232 | 0.8306 | 4.3487 | 0.9069 |
| | CatBoost | 3.1231 | 0.7542 | 3.6227 | 0.8185 | 4.3571 | 0.8958 |
| | LightGBM | 3.2540 | 0.7634 | 3.6211 | 0.8426 | 4.3285 | 0.9144 |
| | LSTM | 3.1773 | 0.8398 | 3.6160 | 0.8958 | 4.1834 | 0.9394 |
| | TCN | 3.2586 | 0.8444 | 3.7447 | 0.8977 | 4.2845 | 0.9454 |
| | MWNet | 3.2312 | 0.8398 | 3.7022 | 0.8869 | 4.2746 | 0.9426 |

|   | Model | | | | | | |
|---|---|---|---|---|---|---|---|
|   | AFE-TFNet | 3.0326 | 0.8452 | 3.5882 | 0.8977 | 4.1735 | 0.9472 |
| B | NaiveDrift | 1.9362 | 0.7047 | 2.1271 | 0.7985 | 2.3692 | 0.8915 |
|   | XGBoost | 1.9149 | 0.7129 | 2.1043 | 0.7985 | 2.3837 | 0.8956 |
|   | CatBoost | 1.9539 | 0.6983 | 2.1371 | 0.7798 | 2.3966 | 0.8777 |
|   | LightGBM | 1.9049 | 0.7358 | 2.0715 | 0.8031 | 2.3474 | 0.9052 |
|   | LSTM | 1.2352 | 0.8452 | 1.3970 | 0.8938 | 1.6586 | 0.9428 |
|   | TCN | 1.2390 | 0.8452 | 1.4008 | 0.8938 | 1.6569 | 0.9396 |
|   | MWNet | 1.2395 | 0.7935 | 1.3312 | 0.8288 | 1.6104 | 0.8901 |
|   | AFE-TFNet | 1.1794 | 0.8608 | 1.3284 | 0.9016 | 1.6055 | 0.9501 |
| C | NaiveDrift | 3.8336 | 0.5842 | 4.3597 | 0.7006 | 5.1381 | 0.8397 |
|   | XGBoost | 3.7762 | 0.5721 | 4.3924 | 0.7083 | 5.1247 | 0.8365 |
|   | CatBoost | 3.7980 | 0.5788 | 4.4031 | 0.7121 | 5.1448 | 0.8424 |
|   | LightGBM | 3.7552 | 0.5621 | 4.3250 | 0.6880 | 5.0781 | 0.8274 |
|   | LSTM | 3.2911 | 0.8465 | 3.7905 | 0.8931 | 4.7631 | 0.9484 |
|   | TCN | 3.4047 | 0.8528 | 3.9027 | 0.8958 | 4.8021 | 0.9475 |
|   | MWNet | 3.3347 | 0.8411 | 3.8770 | 0.8886 | 4.7969 | 0.9357 |
|   | AFE-TFNet | 3.2449 | 0.8611 | 3.7623 | 0.8954 | 4.7326 | 0.9488 |

Fig. 16 illustrates AFE-TFNet's prediction intervals, demonstrating effective coverage of observed wave heights across all confidence levels. The intervals successfully capture wave height variations, including complex peak and trough patterns, with minimal outliers at extreme points. These interval forecasting results further validate AFE-TFNet's robust and comprehensive forecasting capabilities, demonstrating excellence in both point forecasting and uncertainty quantification. This dual capability provides valuable insights into risk assessment and decision-making in wave-energy applications.

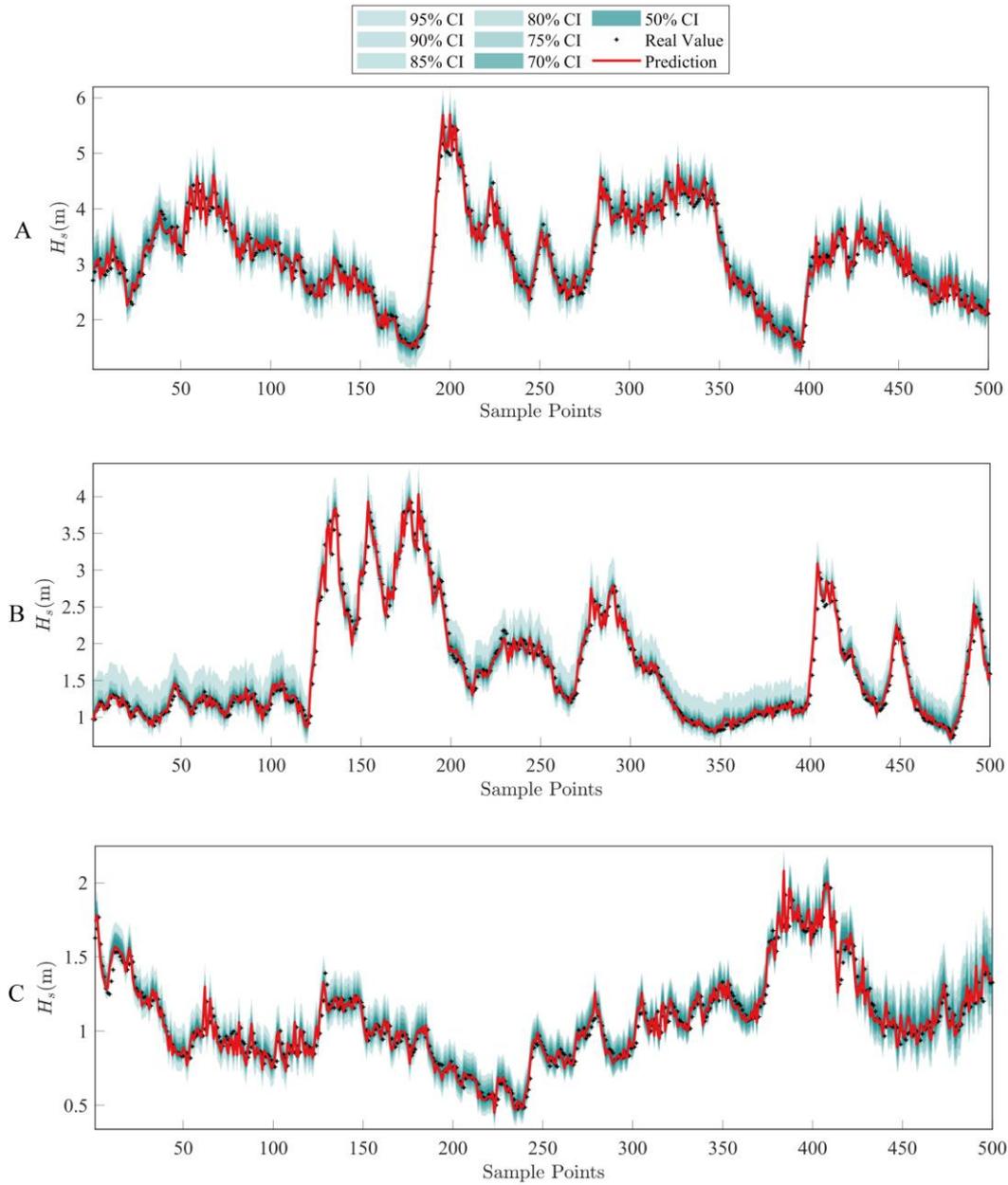

Fig. 16. Interval prediction results.

## 4.6. Ablation experiments

To further verify the effectiveness of the proposed methods, this study conducted a series of ablation experiments by introducing four different models. Each model is designed to exclude or modify key components of the AFE-TFNet architecture, including the feature extraction and energy weighting mechanisms. These models are tested across three datasets to evaluate the impact of each component.

1. wo/fe (without feature extraction): The model without the feature extraction evaluates the impact of feature extraction on performance.
2. w/fft (with FFT): The model with only FFT applied during the feature extraction assesses the effectiveness of FFT.
3. w/wt (with WT): The model with only WT applied during the feature extraction examines the contribution of WT.
4. wo/wei (without weighting): The model without the DHSEW component analyzes its contribution.

### 4.6.1. Impact of feature extraction stage

To validate the effectiveness of the feature extraction stage, Fig. 17 presents bar plots comparing the performance of models with different feature extraction methods. The comparison between AFE-TFNet and the wo/fe model (without feature extraction) shows that omitting this stage significantly reduces performance across all three datasets, particularly for the 6 h and 12 h predictions. Specific metrics in Table 12 show that AFE-TFNet surpasses the wo/fe model by 12.65%, 13.71%, 16.89%, and 2.26% for *RMSE*, *MAE*, *MAPE*, and *R*, respectively. These results underscore the essential role of feature extraction in boosting model performance, especially in scenarios requiring precise long-term forecasting. In comparison with the w/fft and w/wt models, AFE-TFNet consistently outperforms both models in most performance metrics. This suggests that the combination of multiple feature extraction methods offers significant advantages over using FFT or WT alone. Specifically, the integration of FFT and WT allows AFE-TFNet to capture both global and local frequency information more effectively, leading to more accurate predictions, particularly in scenarios with complex data patterns.

To further evaluate the impact of the feature extraction approach, Fig. 18 illustrates scatter plots showing the performance of various ablation models under different prediction scenarios. Scatter points concentrated around the diagonal indicate that the models' predictions are closer to the observed values, reflecting higher prediction accuracy. The results show that AFE-TFNet generally outperforms other models, particularly in long-term forecasts, where this clustering is more pronounced. In contrast, the trendlines of w/fft and w/wt deviate significantly from the diagonal at certain times, especially in the range of larger predicted values. This suggests that these models perform poorly under extreme fluctuations, with more pronounced prediction errors.

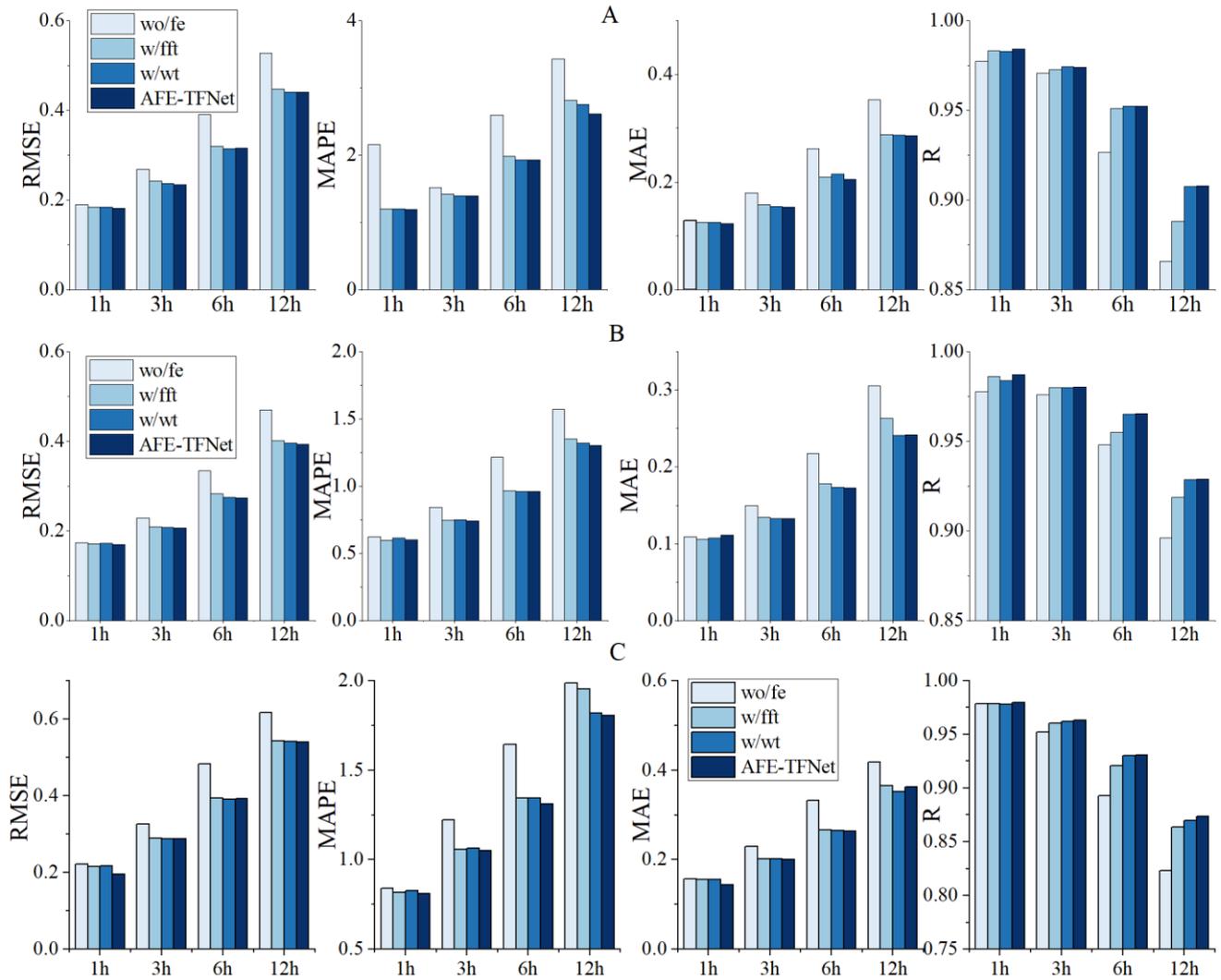

Fig. 17. Bar chart of ablation study results on feature extraction stage.

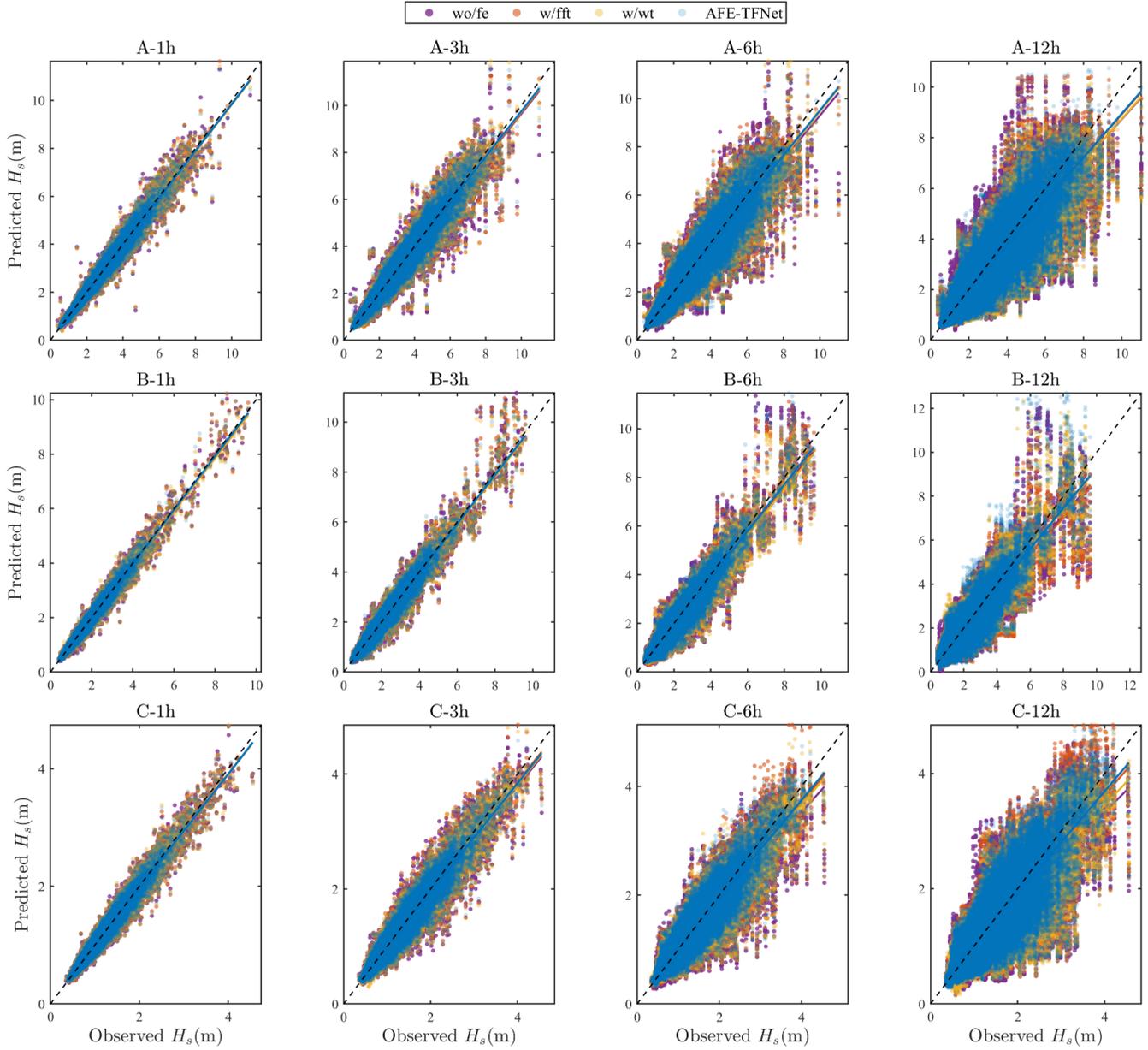

Fig. 18. Scatter chart of ablation study results on feature extraction stage.

Table 12 Results of ablation study on feature extraction stage.

| Station | Model | Metrics | Prediction time | | | |
|---|---|---|---|---|---|---|
| | | | 1h | 3h | 6h | 12h |
| A | wo/fe | RMSE | 0.1889 | 0.2679 | 0.3912 | 0.5285 |
| | | MAE | 0.1277 | 0.1793 | 0.2622 | 0.3535 |
| | | MAPE | 2.1599 | 1.5162 | 2.5949 | 3.4281 |
| | | R | 0.9772 | 0.9709 | 0.9267 | 0.8656 |
| | w/fft | RMSE | 0.1847 | 0.2414 | 0.3204 | 0.4470 |
| | | MAE | 0.1249 | 0.1578 | 0.2094 | 0.2881 |
| | | MAPE | 1.1979 | 1.4200 | 1.9806 | 2.8184 |
| | | R | 0.9834 | 0.9727 | 0.9509 | 0.8880 |
| | w/wt | RMSE | 0.1841 | 0.2368 | 0.3140 | 0.4410 |
| | | MAE | 0.1246 | 0.1542 | 0.2148 | 0.2874 |
| | | MAPE | 1.1973 | 1.3964 | 1.9310 | 2.7542 |
| | | R | 0.9830 | 0.9744 | 0.9523 | 0.9074 |
| | AFE-TFNet | RMSE | 0.1821 | 0.2341 | 0.3156 | 0.4409 |
| | | MAE | 0.1231 | 0.1538 | 0.2054 | 0.2860 |
| | | MAPE | 1.1901 | 1.3927 | 1.9257 | 2.6173 |
| | | R | 0.9842 | 0.9740 | 0.9524 | 0.9078 |
| B | wo/fe | RMSE | 0.1729 | 0.2293 | 0.3347 | 0.4698 |

| Station | Model | Metrics | 1h | 3h | 6h | 12h |
|---|---|---|---|---|---|---|
| | | MAE | 0.1089 | 0.1497 | 0.2169 | 0.3056 |
| | | MAPE | 0.6224 | 0.8431 | 1.2146 | 1.5719 |
| | | R | 0.9777 | 0.9759 | 0.9482 | 0.8960 |
| | w/fft | RMSE | 0.1709 | 0.2093 | 0.2828 | 0.4017 |
| | | MAE | 0.1058 | 0.1345 | 0.1784 | 0.2632 |
| | | MAPE | 0.5940 | 0.7436 | 0.9666 | 1.3532 |
| | | R | 0.9861 | 0.9800 | 0.9549 | 0.9187 |
| | w/wt | RMSE | 0.1725 | 0.2074 | 0.2750 | 0.3957 |
| | | MAE | 0.1071 | 0.1330 | 0.1733 | 0.2413 |
| | | MAPE | 0.6139 | 0.7490 | 0.9622 | 1.3211 |
| | | R | 0.9839 | 0.9800 | 0.9653 | 0.9285 |
| | AFE-TFNet | RMSE | 0.1693 | 0.2071 | 0.2735 | 0.3930 |
| | | MAE | 0.1113 | 0.1328 | 0.1729 | 0.2420 |
| | | MAPE | 0.6025 | 0.7396 | 0.9597 | 1.3027 |
| | | R | 0.9873 | 0.9803 | 0.9655 | 0.9290 |
| C | wo/fe | RMSE | 0.2214 | 0.3265 | 0.4826 | 0.6159 |
| | | MAE | 0.1577 | 0.2300 | 0.3324 | 0.4185 |
| | | MAPE | 0.8386 | 1.2210 | 1.6442 | 1.9864 |
| | | R | 0.9786 | 0.9521 | 0.8929 | 0.8232 |
| | w/fft | RMSE | 0.2161 | 0.2893 | 0.3931 | 0.5426 |
| | | MAE | 0.1560 | 0.2023 | 0.2662 | 0.3653 |
| | | MAPE | 0.8157 | 1.0563 | 1.3453 | 1.9530 |
| | | R | 0.9782 | 0.9603 | 0.9210 | 0.8636 |
| | w/wt | RMSE | 0.2175 | 0.2879 | 0.3910 | 0.5409 |
| | | MAE | 0.1562 | 0.2013 | 0.2656 | 0.3523 |
| | | MAPE | 0.8259 | 1.0638 | 1.3449 | 1.8183 |
| | | R | 0.9780 | 0.9618 | 0.9303 | 0.8697 |
| | AFE-TFNet | RMSE | 0.1962 | 0.2874 | 0.3923 | 0.5405 |
| | | MAE | 0.1435 | 0.2008 | 0.2639 | 0.3623 |
| | | MAPE | 0.8104 | 1.0519 | 1.3094 | 1.8037 |
| | | R | 0.9793 | 0.9630 | 0.9306 | 0.8735 |

### 4.6.2. Impact of DHSEW mechanism

To verify the superiority of DHSEW method, this study compared wo/wei and AFE-TFNet across three datasets. Table 13 and Fig. 19 present the performance metrics of the two models with different prediction times. Specifically, in the 1-h prediction, AFE-TFNet achieved a lower *MAE* than wo/wei, with reductions of approximately 8% at Dataset C. This trend continues in Dataset A and B, although the differences are less pronounced at Dataset A. Both models achieved high accuracy in the 3 h prediction, AFE-TFNet consistently outperformed wo/wei in most short-term forecasts, particularly in metrics such as *MAE* and *MAPE*, where it demonstrated a consistent advantage. As the prediction horizon extended to 6 h and 12 h, the performance of both models degraded, but AFE-TFNet generally demonstrates better accuracy. To visually compare the performance, Fig. 20 compares the absolute cumulative error of wo/wei and AFE-TFNet. In each subplot, the horizontal axis represents the number of samples, while the vertical axis denotes the absolute cumulative error. It can be observed that AFE-TFNet demonstrated a slight but consistent advantage in short-term predictions, while the improvements were more significant in medium- and long-term predictions. This advantage was most noticeable at station C, followed by station A, with station B showing relatively smaller differences. Overall, AFE-TFNet leverages the DHSEW mechanism to dynamically adjust temporal and frequency-domain information, improving its effectiveness in complex forecasting scenarios. Its stability and accuracy are particularly notable in long-term predictions.

Table 13 Results of ablation study on DHSEW.

| Station | Model | Metrics | Prediction time | | | |
|---|---|---|---|---|---|---|
| | | | 1h | 3h | 6h | 12h |
| A | wo/wei | RMSE | 0.1854 | 0.2415 | 0.3243 | 0.4438 |
| | | MAE | 0.1261 | 0.1602 | 0.2115 | 0.2905 |
| | | MAPE | 1.2483 | 1.5115 | 2.0566 | 2.7098 |
| | | R | 0.9838 | 0.9623 | 0.9397 | 0.8948 |
| | AFE-TFNet | RMSE | 0.1821 | 0.2341 | 0.3156 | 0.4409 |
| | | MAE | 0.1231 | 0.1538 | 0.2054 | 0.2860 |

|   |   | MAPE | 1.1901 | 1.3927 | 1.9257 | 2.6173 |
|---|---|---|---|---|---|---|
|   |   | R    | 0.9842 | 0.9740 | 0.9524 | 0.9078 |
| B | wo/wei | RMSE | 0.1638 | 0.2294 | 0.3346 | 0.4697 |
|   |   | MAE  | 0.1288 | 0.1496 | 0.2169 | 0.3055 |
|   |   | MAPE | 0.6223 | 0.8431 | 1.2145 | 1.5718 |
|   |   | R    | 0.9776 | 0.9758 | 0.9482 | 0.8959 |
|   | AFE-TFNet | RMSE | 0.1693 | 0.2071 | 0.2735 | 0.3930 |
|   |   | MAE  | 0.1113 | 0.1328 | 0.1729 | 0.2420 |
|   |   | MAPE | 0.6025 | 0.7396 | 0.9597 | 1.3027 |
|   |   | R    | 0.9873 | 0.9803 | 0.9655 | 0.9290 |
| C | wo/wei | RMSE | 0.2190 | 0.2963 | 0.3995 | 0.5430 |
|   |   | MAE  | 0.1568 | 0.2082 | 0.2716 | 0.3592 |
|   |   | MAPE | 0.8255 | 1.0797 | 1.3485 | 1.7199 |
|   |   | R    | 0.9786 | 0.9606 | 0.9180 | 0.8526 |
|   | AFE-TFNet | RMSE | 0.1962 | 0.2874 | 0.3923 | 0.5405 |
|   |   | MAE  | 0.1435 | 0.2008 | 0.2639 | 0.3623 |
|   |   | MAPE | 0.8104 | 1.0519 | 1.3094 | 1.8037 |
|   |   | R    | 0.9793 | 0.9630 | 0.9306 | 0.8735 |

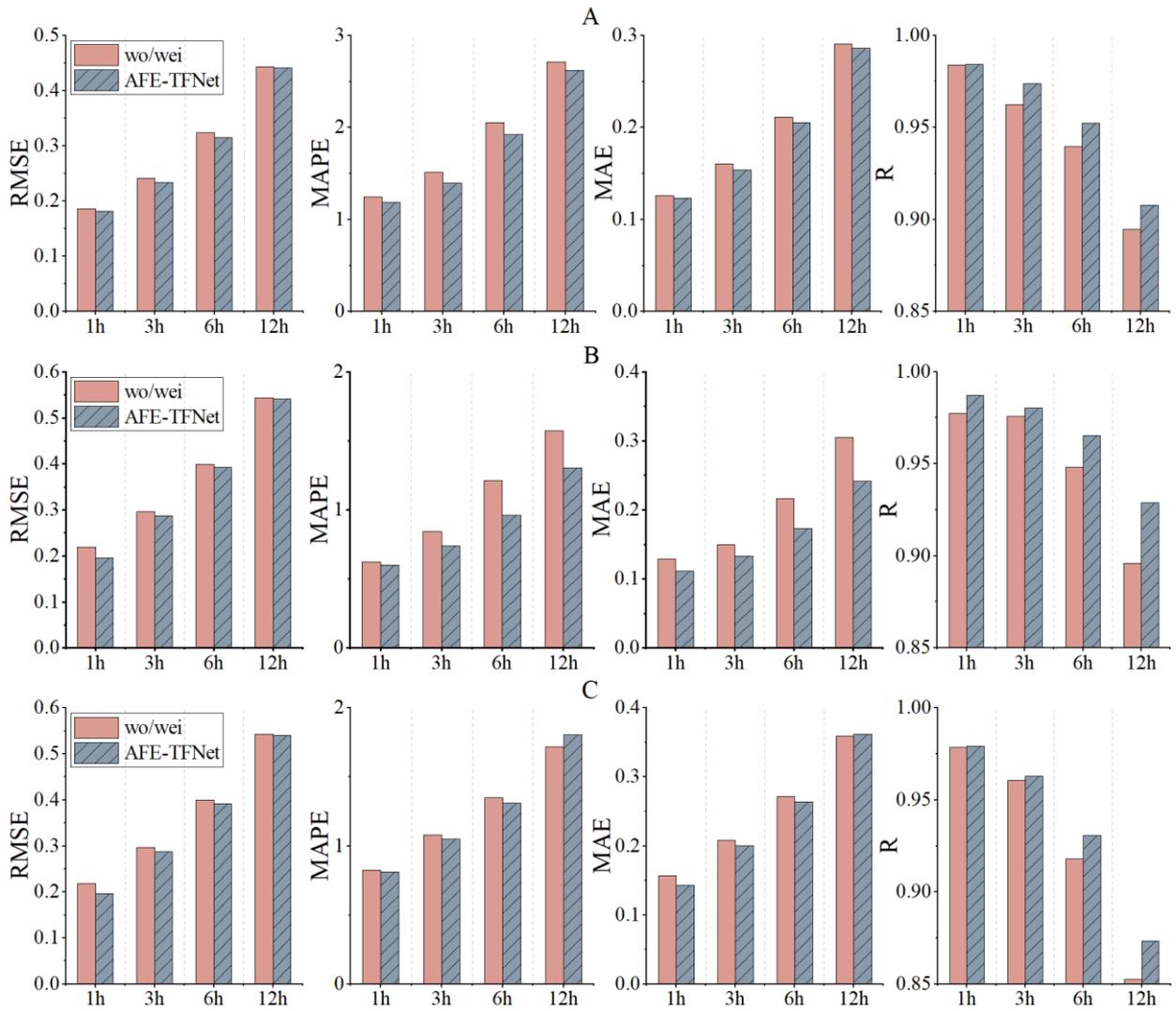

Fig. 19. Bar chart of ablation study results on DHSEW.

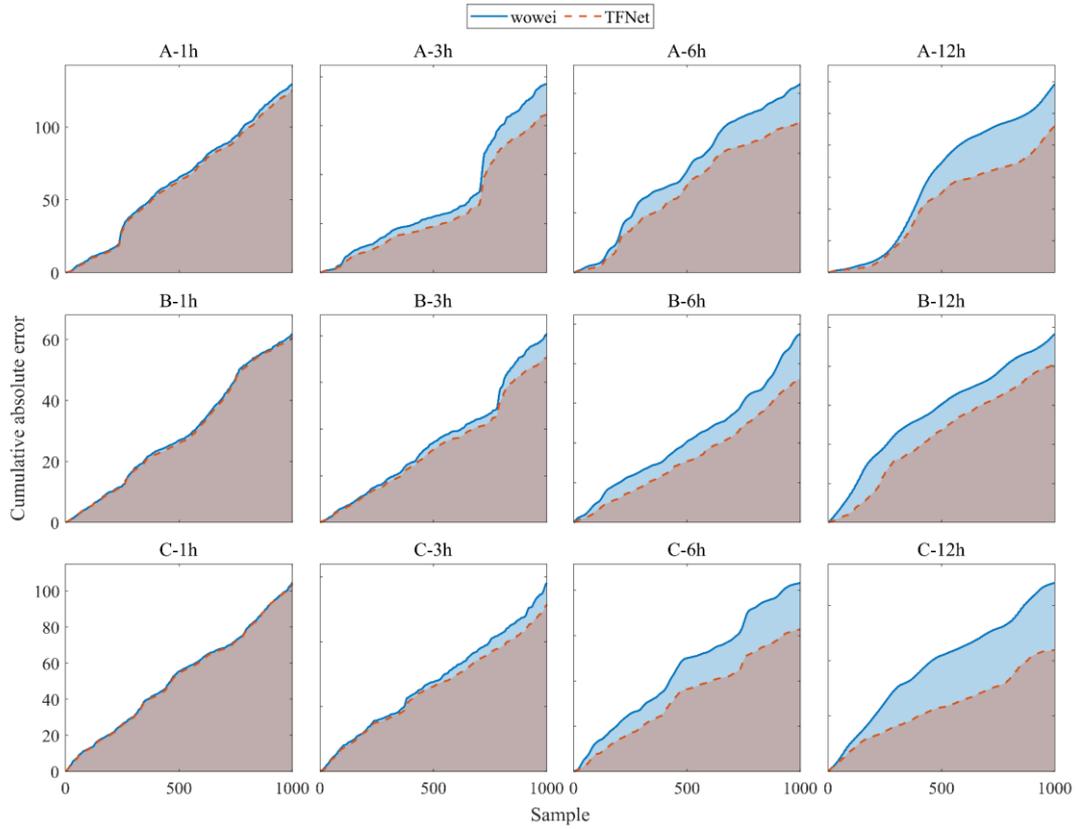

Fig. 20. Absolute cumulative error.

**4.6.3. Sensitivity analysis of rolling time window size**

  To validate the robustness of the proposed model, this study investigates the impact of different sequence length on performance. Sequence lengths of 12, 24, 32, and 40 time steps were tested. The results are shown in Fig. 21, illustrating the changes in *RMSE*, *MAE*, *MAPE*, and *R* for prediction lengths ranging from 1-hour, 3-hour, 6-hour and 12-hour. The results show small fluctuations in *RMSE*, *MAE*, *MAPE*, and *R*. This indicates that the model is nearly insensitive to sequence length, consistently maintaining high predictive performance and exhibiting excellent robustness. Based on these results, a sequence length of 24 time steps is used as the default setting in this study, as it provides a good balance between performance and computational efficiency.

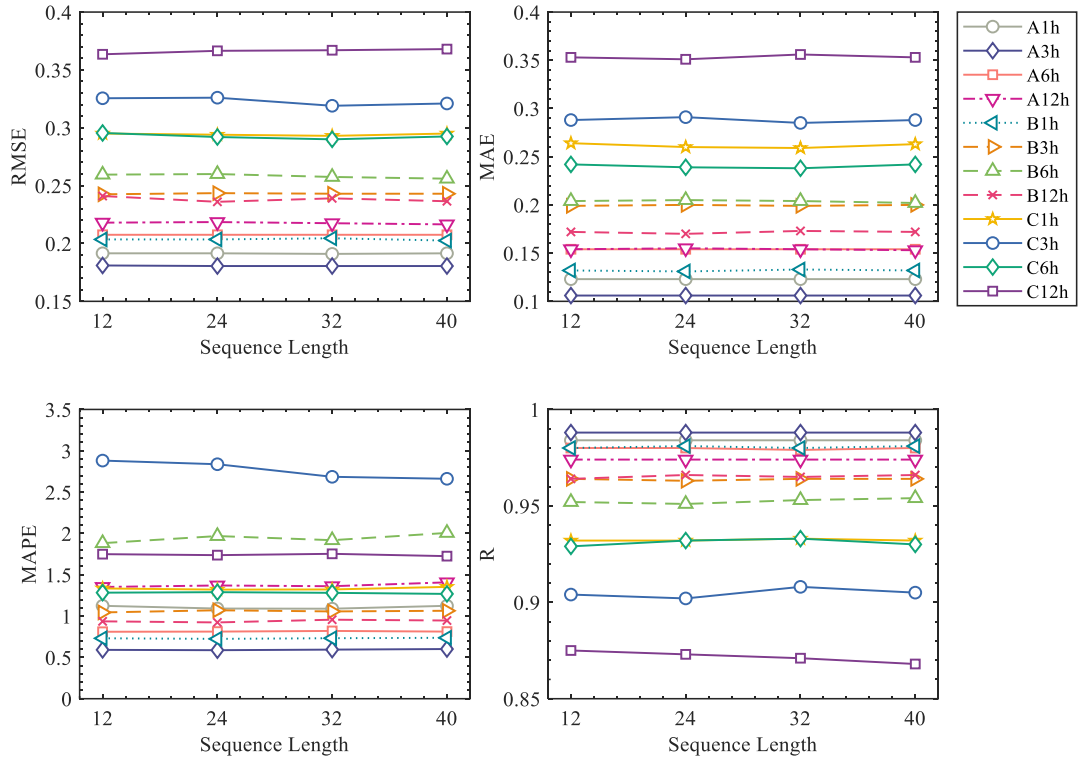

Fig. 21. Effect of sequence length.

# 5. Conclusions

Significant wave height is erratic and random, which makes forecasting more challenging and hinders the development of wave energy. To improve significant wave height prediction, this paper proposes a rolling framework named Adaptive Feature Extraction Time-Frequency Network (AFE-TFNet) which adopts an encoder-decoder architecture. The encoder is designed with two key stages: feature extraction and feature fusion. In the feature extraction stage, Wavelet transform (WT) and Fast Fourier transform (FFT) are combined to capture local and global frequency-domain features. Additionally, an Inception block is used to enhance the extraction of multi-scale information and reducing information loss. In the feature fusion stage, time- and frequency-domain features are integrated using the Dominant Harmonic Sequence Energy Weighting (DHSEW) mechanism. The fused multi-dimensional features are then processed by the long short-term memory (LSTM) network in the decoder. LSTM effectively captures the temporal dependencies in time series, making it well-suited for predicting significant wave height $H_s$. The main conclusions are summarized as follows:

(1) The proposed AFE-TFNet consistently outperforms benchmark models across three datasets, as demonstrated by *RMSE*, *MAE*, *MAPE*, and *R* statistical metrics. While the performance of all models declines as the prediction horizon extends, AFE-TFNet retains a clear advantage, particularly in long-term forecasts. At the 12 h prediction, AFE-TFNet shows the greatest significant performance advantage over the benchmark models, demonstrating its robustness and superior ability to predict significant wave heights across diverse datasets.

(2) The ablation studies validate the effectiveness of each component (wo/fe, w/fft, w/wt, wo/wei) of AFE-TFNet. Models with the feature extraction stage significantly outperform those without it, achieving average improvements of 12.65%, 13.71%, 16.89%, and 2.26% in *RMSE*, *MAE*, *MAPE*, and *R*, respectively. This improvement can be attributed to the model's ability to capture multi-frequency components more effectively, highlighting the importance of combining both global and local frequency information to enhance prediction accuracy.

(3) A comprehensive comparison between AFE-TFNet and the model without the DHSEW mechanism across three datasets reveals that AFE-TFNet demonstrates better stability and accuracy, particularly in medium- and long-term predictions. The absolute cumulative error analysis confirms AFE-TFNet's advantage in handling complex predictions by effectively capturing both time- and frequency-domain information, improving its ability to identify patterns and trends in short- and long-term forecasts.

In this study, AFE-TFNet has shown promising results in improving significant wave height prediction. Its ability to integrate multi-scale frequency- and time-domain information, along with its effectiveness in handling complex non-linear and non-stationary signals. This framework shows strong potential as an effective tool for enhancing significant wave height forecast accuracy, thereby supporting more reliable decision-making in wave energy development and utilization. Future research should consider integrating additional cutting-edge multiscale analysis techniques and deep learning architectures to further refine predictive performance. Additionally, a comprehensive evaluation system is needed to assess the influence of various environmental conditions on prediction outcomes, with a focus on different temporal and spatial scales. Such efforts would help optimize the model's adaptability and robustness across diverse scenarios.